# An Introduction to the Practical and Theoretical Aspects of Mixture-of-Experts Modeling

Hien D. Nguyen* and Faicel Chamroukhi†

July 11, 2017

**Abstract**

Mixture-of-experts (MoE) models are a powerful paradigm for modeling of data arising from complex data generating processes (DGPs). In this article, we demonstrate how different MoE models can be constructed to approximate the underlying DGPs of arbitrary types of data. Due to the probabilistic nature of MoE models, we propose the maximum quasi-likelihood (MQL) estimator as a method for estimating MoE model parameters from data, and we provide conditions under which MQL estimators are consistent and asymptotically normal. The blockwise minorization–maximizatoin (blockwise-MM) algorithm framework is proposed as an all-purpose method for constructing algorithms for obtaining MQL estimators. An example derivation of a blockwise-MM algorithm is provided. We then present a method for constructing information criteria for estimating the number of components in MoE models and provide justification for the classic Bayesian information criterion (BIC). We explain how MoE models can be used to conduct classification, clustering, and regression and we illustrate these applications via a pair of worked examples.

*Department of Mathematics and Statistics, La Trobe University, 3086 Bundoora, Victoria Australia. Corresponding author. Email: h.nguyen5@latrobe.edu.au.

†Normandie Universite, UNICAEN, CNRS, Laboratoire de Mathematiques Nicolas Oresme - LMNO, 14000 Caen, France.

# 1 Introduction

Let $\boldsymbol{D}^\top = \left(\boldsymbol{X}^\top, \boldsymbol{Y}^\top\right) \in \mathbb{X} \times \mathbb{Y}$ be an observed random pair from some data generating process (DGP), where $\mathbb{X} \subset \mathbb{R}^p$ and $\mathbb{Y} \subset \mathbb{R}^q$ for $p, q \in \mathbb{N}$. Here, $(\cdot)^\top$ is the matrix transposition operator. We shall call $\boldsymbol{X}$ the input variable and $\boldsymbol{Y}$ the output (or response) variable. Suppose that the DGP of interest can be approximated as follows.

Firstly, suppose that there is an unobserved random variable $Z \in [g] = \{1, \dots, g\}$ ($g \in \mathbb{N}$), where by the conditional relationship between $Z$ and the input can be characterized by

$$\mathbb{P}\left(Z = z | \boldsymbol{X} = \boldsymbol{x}\right) = \text{Gate}_z\left(\boldsymbol{x}; \boldsymbol{\gamma}\right), \tag{1}$$

where $\boldsymbol{\gamma} \in \mathbb{R}^{d_\gamma}$ ($d_\gamma \in \mathbb{N}$) is some parameter vector, $\text{Gate}_z\left(\boldsymbol{x}; \boldsymbol{\gamma}\right) > 0$, and $\sum_{z=1}^{g} \text{Gate}_z\left(\boldsymbol{x}; \boldsymbol{\gamma}\right) = 1$. Secondly, let the conditional relationship between the response and the input, given $Z = z$, be characterized by

$$f\left(\boldsymbol{y} | \boldsymbol{X} = \boldsymbol{x}, Z = z\right) = \text{Expert}_z\left(\boldsymbol{y} | \boldsymbol{x}; \boldsymbol{\eta}_z\right), \tag{2}$$

where $\boldsymbol{\eta}_z \in \mathbb{R}^{d_\eta}$ ($d_\eta \in \mathbb{N}$) is some parameter vector and $\text{Expert}_z\left(\boldsymbol{y} | \boldsymbol{x}; \boldsymbol{\eta}_z\right)$ is a probability density function or probability mass function (PDF or PMF; see e.g. DasGupta, 2011, Chs. 2 and 3). Via characterizations (1) and (2), and using the law of total probability, we can characterize the marginal relationship between the response and the input, unconditional on $Z$, via the expression

$$\mathrm{MoE}\left(\boldsymbol{y}|\boldsymbol{x};\boldsymbol{\theta}\right)=\sum_{z=1}^{g}\mathrm{Gate}_{z}\left(\boldsymbol{x};\boldsymbol{\gamma}\right)\mathrm{Expert}_{z}\left(\boldsymbol{y}|\boldsymbol{x};\boldsymbol{\eta}_{z}\right) \tag{3}$$
$$=f\left(\boldsymbol{y}|\boldsymbol{X}=\boldsymbol{x}\right),$$

where $f\left(\boldsymbol{y}|\boldsymbol{X}=\boldsymbol{x}\right)$ is a PDF of the response $\boldsymbol{y}$ given the input $\boldsymbol{X}=\boldsymbol{x}$. Here, $\boldsymbol{\theta}^{\top}=\left(\boldsymbol{\gamma}^{\top},\boldsymbol{\eta}_{1}^{\top},\ldots,\boldsymbol{\eta}_{g}^{\top}\right)$ is the vector of all parameter elements that are required in characterizing (3). We refer to the approximation of the DGP of $\boldsymbol{D}$ of form (3) as a $g$-component mixture-of-experts (MoE) model. The functions $\mathrm{Gate}_z$ and $\mathrm{Expert}_z$ are referred to as gating and expert functions, respectively.

MoE models were first studied as neural networks (NNs) by Jacobs et al. (1991), where they were used to model complex and heterogeneous DGPs. A schematic diagram of an MoE model as a NN is provided in Figure 1. Some recent reviews of the MoE literature are provided by Yuksel et al. (2012) and Masoudnia & Ebrahimpour (2014).

MoE models have been broadly applied to numerous areas of business, science, and technology for the tasks of classification, clustering, and regression. A sample of recent applications that were not covered by Yuksel et al. (2012) and Masoudnia & Ebrahimpour (2014) includes: modeling neural connectivity (Bock & Fine, 2014), fusion and segmentation of images (Camplani et al., 2014), segmentation of spectral images (Cohen & Le Pennec, 2014), phone activity recognition (Lee & Cho, 2014), climatic change modeling (Nguyen & McLachlan, 2014), parallel mapping of threads in dynamic runtime environments (Emani & O'Boyle, 2015), cardiac stress monitoring via heart sounds (Herzig et al., 2015), aerodynamic performance predictions (Liem et al., 2015), functional magnetic resonance image analysis (Shoenmakers et al., 2015), heterogeneity modeling in neural connectivity data (Eavani et al., 2016), reinforcement learning (He et al., 2016), landmine detection (Yuksel & Gader, 2016), and attention deficit hyperactivity disorder diagnosis (ADHD) (Yaghoobi Karimu & Azadi, 2017).

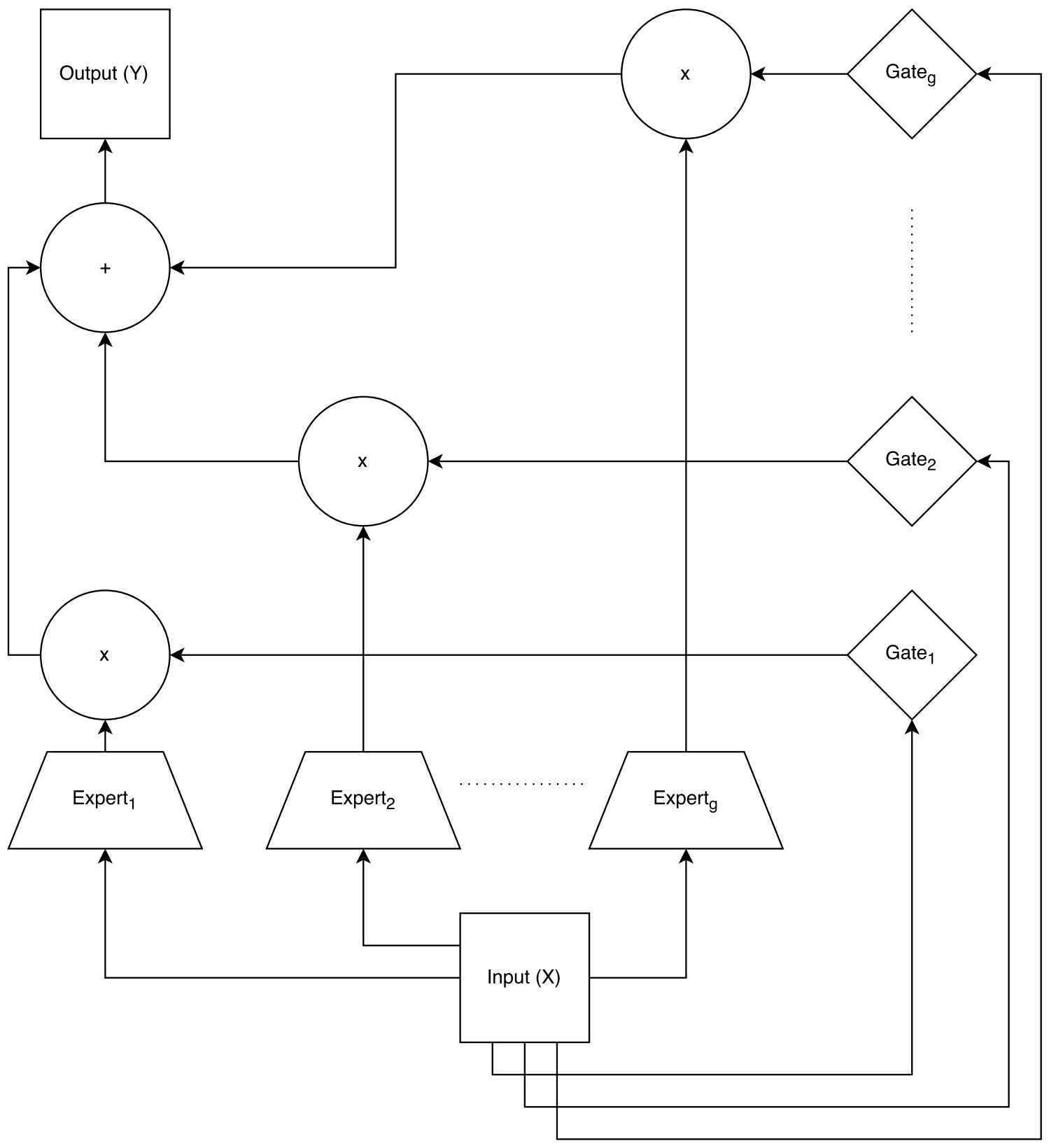

Figure 1: Schematic diagram of the NN architecture of a $g$-component MoE model as defined by characterizations (1), (2), and (3).

Since the reviews by Yuksel et al. (2012) and Masoudnia & Ebrahimpour (2014), there have also been numerous theoretical developments regarding the approximation capacity of MoE models (Mendes & Jiang, 2012; Norets & Pelenis, 2014; Nguyen et al., 2016), the performance of maximum quasi-likelihood (MQL) estimation algorithms and the properties of MQL estimators (Nguyen & McLachlan, 2014, 2016), and the modes by which model selection can be conducted within the MoE framework (Montuelle & Le Pennec, 2014; Baudry, 2015).

The goal of this article is to provide a concise treatment regarding the practice of constructing MoE models as well as the theoretical justification of such models. As such, the remainder of the article progresses as follows. In Section 2, we discuss the construction of MoE models via the choice of gating and expert functions. In Section 3, we present some of the aforementioned recent theoretical results in a digestible manner and demonstrate their use, where possible. In Section 4, we discuss the problems of classification, clustering, and regression, and show how MoE models can be applied to each of these tasks. In Section 5, we provide some examples of each task. Conclusions are finally drawn in Section 6.

# 2 Mixture-of-Experts Modeling

We begin by considering the original MoE model of Jacobs et al. (1991) that was designed for the task of multi-speaker vowel discrimination. Here, Jacobs et al. (1991) paired the popular and ubiquitous soft-max gating function

$$\text{Gate}_z(\boldsymbol{x};\boldsymbol{\gamma}) = \frac{\exp\left(\alpha_{z0} + \boldsymbol{\alpha}_z^\top \boldsymbol{x}\right)}{\sum_{\zeta=1}^{g} \exp\left(\alpha_{\zeta 0} + \boldsymbol{\alpha}_\zeta^\top \boldsymbol{x}\right)}, \tag{4}$$

with the multivariate Gaussian distribution expert

$$\text{Expert}_z(\boldsymbol{y}|\boldsymbol{x};\boldsymbol{\eta}_z) = \phi(\boldsymbol{y};\boldsymbol{\mu}_z,\boldsymbol{\Sigma}_z), \tag{5}$$

where

$$\phi(\boldsymbol{y};\boldsymbol{\mu},\boldsymbol{\Sigma}) = |2\pi\boldsymbol{\Sigma}|^{-1/2} \exp\left[-\frac{1}{2}(\boldsymbol{y}-\boldsymbol{\mu})^\top \boldsymbol{\Sigma}^{-1}(\boldsymbol{y}-\boldsymbol{\mu})\right]$$

is the multivariate normal density function with mean vector $\boldsymbol{\mu} \in \mathbb{R}^q$ and positive-definite covariance matrix $\boldsymbol{\Sigma} \in \mathbb{R}^{q\times q}$. Here $\alpha_{z0} \in \mathbb{R}$ and $\boldsymbol{\alpha}_z^\top = (\alpha_{z1},\dots,\alpha_{zp}) \in \mathbb{R}^p$ for each $z \in [g-1]$, and $\alpha_{g0} = 0$ and $\boldsymbol{\alpha}_g = \mathbf{0}$, where $\mathbf{0}$ is the zero vector. We set $\boldsymbol{\gamma}^\top = \left(\alpha_{10}, \boldsymbol{\alpha}_1^\top, \dots, \alpha_{g-1,0}, \boldsymbol{\alpha}_{g-1}^\top\right)$ and $\boldsymbol{\eta}_z^\top = \left(\boldsymbol{\mu}_z^\top, \text{vech}^\top \boldsymbol{\Sigma}_z\right)$, where $\text{vech}(\cdot)$ extracts the unique elements of a symmetric matrix (cf. Henderson & Searle, 1979). The MoE of Jacobs et al. (1991) is a direct extension of the usual Gaussian mixture model (GMM; see e.g. McLachlan & Peel, 2000, Ch. 3) that allows for mixing proportions (i.e. the probabilities of $Z = z$, for each $z \in [g]$) to depend on the input variable.

In Jordan & Jacobs (1994), the expert of form (5) was extended upon via the multivariate Gaussian regression expert

$$\text{Expert}_z(\boldsymbol{y}|\boldsymbol{x};\boldsymbol{\eta}_z) = \phi(\boldsymbol{y};\boldsymbol{b}_z + \mathbf{B}_z\boldsymbol{x},\boldsymbol{\Sigma}_z), \tag{6}$$

where $\boldsymbol{b}_z \in \mathbb{R}^q$ and $\mathbf{B}_z \in \mathbb{R}^{q\times p}$, for each $z \in [g]$. Here, $\boldsymbol{\eta}_z^\top = \left(\boldsymbol{b}_z^\top, \text{vec}^\top \mathbf{B}_z\right)$, where $\text{vec}(\cdot)$ puts all elements of the matrix input into a vector (cf. Henderson & Searle, 1979). Whereas the MoE model with expert (5) can be seen as an extension of the GMM, the MoE with expert (6) is analogously an extension of the multivariate Gaussian mixture regression model of Jones & McLachlan (1992). Robust experts for heterogeneous linear regression models have also been considered by Nguyen & McLachlan (2016), Chamroukhi (2016), and Chamroukhi (2017), where Laplace, Student-$t$, and skew Student-$t$ experts are used in place of (6), respectively.

## 2.1 Mixture of Generalized Linear Experts Models

In the same way that the MoE model with experts in the form of (6) can be seen as a heterogeneous linear regression model, heterogeneous versions of other generalized linear models (GLMs; cf. Nelder & Wedderburn, 1972 and McCullagh & Nelder, 1989) can be constructed via MoE modeling. The first of such models was considered by Jordan & Jacobs (1994), where the logistic regression expert

$$\text{Expert}_z\left(y|\boldsymbol{x};\boldsymbol{\eta}_z\right)=\left[\frac{\exp\left(\beta_{0z}+\boldsymbol{\beta}_z^\top\boldsymbol{x}\right)}{1+\exp\left(\beta_{0z}+\boldsymbol{\beta}_z^\top\boldsymbol{x}\right)}\right]^y\left[\frac{1}{1+\exp\left(\beta_{0z}+\boldsymbol{\beta}_z^\top\boldsymbol{x}\right)}\right]^{1-y} \tag{7}$$

was proposed for the modeling of binary response data $Y\in\{0,1\}$. Here, $\beta_{0z}\in\mathbb{R}$, $\boldsymbol{\beta}_z\in\mathbb{R}^p$ and $\boldsymbol{\eta}_z^\top=\left(\beta_{z0},\boldsymbol{\beta}_z^\top\right)$ for each $z\in[g]$.

Another MoE model in this class is the MoE with Poisson regression experts of Grun & Leisch (2008), where

$$\text{Expert}_z\left(y|\boldsymbol{x};\boldsymbol{\eta}_z\right)=\frac{\exp\left[y\left(\beta_{0z}+\boldsymbol{\beta}_z^\top\boldsymbol{x}\right)\right]}{y!}\exp\left[-\exp\left(\beta_{0z}+\boldsymbol{\beta}_z^\top\boldsymbol{x}\right)\right],$$

which was proposed for modeling of count response data $Y\in\{0\}\cup\mathbb{N}$. Here, $\boldsymbol{\eta}_z$ is the same as that of (7). Further models considered in the literature include the MoE models with gamma experts of Jiang & Tanner (1999a) for modeling positive responses $Y\in[0,\infty)$ as well as the MoE models with multinomial logistic experts of Chen et al. (1999) for modeling categorical responses $Y\in[K]$ for some $K \in \mathbb{N}$, where $K > 2$.

### 2.2 Gating Functions

The majority of MoE models that are applied in practice tend to utilize soft-max gating functions of form (4). In Xu et al. (1995), the Gaussian gating functions of form

$$\text{Gate}_z\left(\boldsymbol{x};\boldsymbol{\gamma}\right) = \frac{\pi_z \phi\left(\boldsymbol{x};\boldsymbol{m}_z,\boldsymbol{S}_z\right)}{\sum_{\zeta=1}^{g} \pi_\zeta \phi\left(\boldsymbol{x};\boldsymbol{m}_\zeta,\boldsymbol{S}_\zeta\right)} \tag{8}$$

was proposed, where $\pi_z > 0$ for each $z \in [g]$, $\sum_{z=1}^{g} \pi_z = 1$, and

$$\boldsymbol{\gamma}^\top = \left(\pi_1, \boldsymbol{m}_1^\top, \text{vech}^\top \boldsymbol{S}_1, \dots, \pi_g, \boldsymbol{m}_g^\top, \text{vech}^\top \boldsymbol{S}_g\right).$$

The Gating functions of form (8) have seen interest use in the literature under the cluster-weighted modeling framework of Ingrassia et al. (2012) and the Gaussian locally-linear mapping framework of Deleforge et al. (2015). It can be shown that under some restrictions, there is an equivalence between the class of gating functions of form (4) and (8) (cf. Ingrassia et al., 2012 and Norets & Pelenis, 2014).

Although it is possible to utilize any set of functions that meet the restrictions $\text{Gate}_z\left(\boldsymbol{x};\boldsymbol{\gamma}\right) > 0$, and $\sum_{z=1}^{g} \text{Gate}_z\left(\boldsymbol{x};\boldsymbol{\gamma}\right) = 1$, there are few alternatives to (4) and (8) that are considered in the literature. Some of these considered alternatives include the exponential family gating functions of Xu et al. (1995) and the Student-$t$ gating functions of Perthame et al. (2016).

### 2.3 Additional Notes

Aside from the simple MoE models that can be characterized via the simple architecture of Figure 1, there are more intricate constructions that are possible for the modeling of complex data. Examples of extensions to the MoE modeling framework include the Mixed-effects MoE models of Ng & McLachlan (2007) and Ng & McLachlan (2014), and the hierarchical MoE models of Jordan & Jacobs (1992) and Jordan & Jacobs (1994) that can be used to fit highly heterogeneous and nonlinear data. We find the hierarchical MoE models to be particularly interesting moving forward as they present a direction for construction of deep generative NNs. Works in this direction include van den Oord & Schrauwen (2014), Theis & Bethge (2015), and Variani et al. (2015).

# 3 Theoretical Results

## 3.1 Approximation Theorems

We begin by considering some approximation theory results regarding the most popular class of MoE models: the mixture of linear experts with gates of form (4), and

$$\text{Expert}_z\left(y|\boldsymbol{x};\boldsymbol{\eta}_z\right)=h_z\left(y;\beta_{0z}+\boldsymbol{\beta}_z^\top\boldsymbol{x}\right),$$

where $h_z\left(\cdot;\mu\right)$ is a PDF with support $\mathbb{R}$ and mean value $\mu\in\mathbb{R}$. Here, $\boldsymbol{\eta}_z$ is the same as that of (7). Examples of such MoE models include the $q=1$ case of the linear experts of Jordan & Jacobs (1994), the Laplace experts of Nguyen & McLachlan (2016), and the Student-$t$ experts of Chamroukhi (2016). Under characterization (3), the expectation of the response given the input of such MoE models can be written as

$$\begin{aligned}\mathbb{E}\left(Y|\boldsymbol{X}=\boldsymbol{x}\right)&=\sum_{z=1}^{g}\frac{\exp\left(\alpha_{z0}+\boldsymbol{\alpha}_z^\top\boldsymbol{x}\right)}{\sum_{\zeta=1}^{g}\exp\left(\alpha_{\zeta0}+\boldsymbol{\alpha}_\zeta^\top\boldsymbol{x}\right)}\left(\beta_{0z}+\boldsymbol{\beta}_z^\top\boldsymbol{x}\right)\\&=m\left(\boldsymbol{x}\right).\end{aligned}\tag{9}$$

Let $\mathbb{C}\left(\mathbb{X}\right)$ be the class of continuous functions and let

$$\mathbb{M}\left(\mathbb{X}\right)=\left\{m\left(\boldsymbol{x}\right):m\text{ has form (9)},\,\boldsymbol{\theta}\in\mathbb{R}^{2g(p+1)-p-1},\,g\in\mathbb{N}\right\}$$

be the class of mean functions obtained from the mixture of linear experts models described above, over the domain $\mathbb{X}$. The following result from Nguyen et al. (2016) was obtained as a direct consequence of the Stone-Weierstrass theorem (cf. Cotter, 1990).

**Theorem 1.** *If $\mathbb{X}\subset\mathbb{R}^{q}$ is compact, then the class $\mathbb{M}\left(\mathbb{X}\right)$ is dense within the class $\mathbb{C}\left(\mathbb{X}\right)$. That is, for any $c\in\mathbb{C}\left(\mathbb{X}\right)$ and $\epsilon>0$, there exists an $m\in\mathbb{M}\left(\mathbb{X}\right)$ such that $\sup_{\boldsymbol{x}\in\mathbb{X}}\left|c\left(\boldsymbol{x}\right)-m\left(\boldsymbol{x}\right)\right|<\epsilon$.*

Theorem 1 can be viewed as a universal approximation theorem in the style of the famous result by Cybenko (1989). The theorem states that any continuous function over a compact subset of the Euclidean space can be modeled arbitrarily closely by a mixture of linear experts mean function of form (9). Unfortunately, the theorem does not provide an approximation rate.

Let $\left\Vert \cdot\right\Vert _{\mathbb{X},s}$ denote the $\mathcal{L}_{s}$ norm over the support $\mathbb{X}$, for $s\in(1,\infty]$. Define $\mathbb{W}_{s}^{k}\left(\mathbb{X}\right)$ to be the Sobolev class of continuously differentiable functions in $\mathcal{L}^{s}\left(\mathbb{X}\right)$ (i.e. functions with finite $\mathcal{L}_{s}$ norm over the support $\mathbb{X}$) with $k\in\mathbb{N}$ derivatives, where the sum of the $\mathcal{L}_{s}$ norms of the derivatives is bounded. The following result regarding the estimation of functions in Sobolev classes was obtained by Zeevi et al. (1998).

**Theorem 2.** *Assume that $\mathbb{X}\subset\mathbb{R}^{q}$ is compact and define $\mathbb{M}_{g}\left(\mathbb{X}\right)$ to be the subset of $\mathbb{M}\left(\mathbb{X}\right)$ where $g$ is fixed. There exists an absolute positive constant $c$ such that*

$$\sup_{w\in\mathbb{W}_{s}^{k}\left(\mathbb{X}\right)}\inf_{m_{g}\in\mathbb{M}_{g}\left(\mathbb{X}\right)}\left\Vert w\left(\boldsymbol{x}\right)-m_{g}\left(\boldsymbol{x}\right)\right\Vert _{\mathbb{X},s}\le\frac{c}{g^{k/q}}.$$

Theorem 2 sacrifices the generality of approximating over all continuous functions as a tradeoff for a uniform approximation rate result. The theorem states that an increase in the number of components in the MoE model increases the accuracy of approximation. However, the rate of increase is itself accelerated by greater differentiability of the target class and decelerated by increasing dimensionality of the support $\mathbb{X}$.

We now suppose that $\boldsymbol{D} \in \mathbb{R}^p \times \mathbb{R}$ is generated from some unknown DGP that can be characterized by a joint density function $f_0(\boldsymbol{d})$ and where the conditional relationship between the response and the input can be characterized by a conditional density function $f_0(\boldsymbol{y}|\boldsymbol{X}=\boldsymbol{x})$. Suppose that we wish to approximate $f_0(\boldsymbol{y}|\boldsymbol{X}=\boldsymbol{x})$ by an MoE with Gaussian gating functions of form

$$\mathrm{Gate}_z(\boldsymbol{x};\boldsymbol{\gamma}) = \frac{\pi_z \phi\left(\boldsymbol{x};\boldsymbol{m}_z, s_z^2 \mathbf{I}\right)}{\sum_{\zeta=1}^{g} \pi_\zeta \phi\left(\boldsymbol{x};\boldsymbol{m}_\zeta, s_\zeta^2 \mathbf{I}\right)}, \tag{10}$$

where $s_z^2 > 0$ and $\mathbf{I}$ is the identity function, and experts of form

$$\mathrm{Expert}_z(y|\boldsymbol{x};\boldsymbol{\eta}_z) = h(y;\mu_z,\sigma_z), \tag{11}$$

with $\mu_z \in \mathbb{R}$ and $\sigma_z$, for each $z \in [g]$. Here, the function $h\left(\cdot,\mu,\sigma^2\right)$ is taken to be of the form

$$h\left(\cdot;\mu,\sigma^2\right) = \sigma^{-1}\psi\left(\frac{\cdot-\mu}{\sigma}\right),$$

where $\psi(y)$ is a probability density function that is a bounded, continuous and symmetric function (about zero), and monotonically decreasing in $|y|$. Furthermore, we assume that $\log h(y;\mu,\sigma)$ is integrable with respect to the $f_0(\boldsymbol{d})\,\mathrm{d}\boldsymbol{d}$.

Let the class of $g$-component MoE models of form (3) with gating functions of form (10) and experts of form (11) over the support $\mathbb{X}$ be denoted $\mathbb{ME}_g(\mathbb{X})$. Further, denote the Euclidean norm by $\|\cdot\|$. The following result is available from Norets & Pelenis (2014).

**Theorem 3.** *Assume that (A1) $\mathbb{X} \subset \mathbb{R}^q$ is compact, (A2) $f_0(y|\boldsymbol{X}=\boldsymbol{x})$ is*

*continuous in $\boldsymbol{d}$, and (A3) there exists an $s > 0$ such that*

$$\int \log \frac{f_0\left(y|\boldsymbol{X}=\boldsymbol{x}\right)}{\inf_{\|a-y\|\leq s} \inf_{\|\boldsymbol{b}-\boldsymbol{x}\|\leq s} \ f_0\left(a|\boldsymbol{X}=\boldsymbol{b}\right)} f_0\left(\boldsymbol{d}\right) d\boldsymbol{d} < \infty.$$

*If Assumptions (A1)–(A3) are fulfilled then for any $\epsilon > 0$, there exists a $g \in \mathbb{N}$ and an $MoE(y|\boldsymbol{x};\boldsymbol{\theta}) \in \mathbb{ME}_g(\mathbb{X})$, such that*

$$\int \log \frac{f_0\left(y|\boldsymbol{X}=\boldsymbol{x}\right)}{MoE\left(y|\boldsymbol{x};\boldsymbol{\theta}\right)} f_0\left(\boldsymbol{d}\right) d\boldsymbol{d} < \epsilon.$$

Theorem 3 states that the class of MoE models with Gaussian gating and a suitable location-scale experts can densely approximate arbitrary continuous PDFs over compact supports with respect to the conditional Kullback-Leibler divergence (Kullback & Leibler, 1951). This is a powerful result and extends upon well-known denseness theorems regarding approximations of marginal distributions by mixtures of location-scale PDFs (e.g. DasGupta, 2008, Thm. 33.2).

Although Theorem 3 uses Gaussian gating functions, because of the mapping between Gaussian gating and soft-max gating functions, it also applies when the gating functions are of form (4). An alternative denseness result to Theorem 3 is that of Jiang & Tanner (1999a), which is difficult to state but can also be applied to MoE models with GLM experts.

## 3.2 Maximum Quasi-Likelihood Estimation

Let $\{\boldsymbol{D}_i\}_{i=1}^n$ be an IID (independent and identically distributed) random sample of $n \in \mathbb{N}$ observations from some DGP that can be characterized by the PDF $f_0(\boldsymbol{d})$ and where the conditional relationship between each response $\boldsymbol{Y}_i$ given input $\boldsymbol{X}_i$ ($i \in [n]$) can be characterized by the conditional PDF $f_0(\boldsymbol{y}|\boldsymbol{X}=\boldsymbol{x})$. Further, let $\{\boldsymbol{d}_i\}_{i=1}^n$ be some fixed observation of $\{\boldsymbol{D}_i\}_{i=1}^n$. For some fixed $g \in \mathbb{N}$, and without knowledge of either $f_0(\boldsymbol{d})$ and $f_0(\boldsymbol{y}|\boldsymbol{X}=\boldsymbol{x})$, suppose that we wish to estimate the MoE model of form (3), for some class of gating and expert functions, that best approximates $f_0(\boldsymbol{y}|\boldsymbol{X}=\boldsymbol{x})$. As proposed by Zeevi et al. (1998) (see also White, 1982), we can do so by obtaining the MQL estimator $\hat{\boldsymbol{\theta}}_n$: a local maximum of the log-quasi-likelihood function

$$Q_n(\boldsymbol{\theta}) = \sum_{i=1}^{n} \log \mathrm{MoE}(\boldsymbol{y}_i|\boldsymbol{x}_i;\boldsymbol{\theta}). \tag{12}$$

Often, the task of obtaining a local maximum of (12) can be difficult. For example, we cannot obtain closed form solutions to the usual first-order condition (FOC) for differentiable functions when the gating functions are of form (4) and the experts are of form (6). As such, iterative or numerical schemes are often employed to conduct maximization. In Nguyen & McLachlan (2014) and Nguyen & McLachlan (2016), the authors considered the blockwise-MM (minorization–maximization) algorithm framework of Lange (2016); see Nguyen, 2017 for a concise tutorial on MM algorithms.

## 3.3 Minorization–Maximization Algorithms

The blockwise-MM algorithm framework can be described as follows. Suppose that we have some objective function $O(\boldsymbol{u})$, where

$$\boldsymbol{u}^\top = \left(\boldsymbol{u}_1^\top, \dots, \boldsymbol{u}_k^\top\right) \in \mathbb{U} = \prod_{j=1}^{k} \mathbb{U}_j \subset \prod_{j=1}^{k} \mathbb{R}^{d_k}$$

for some $k \in \mathbb{N}$, where $O(\boldsymbol{u})$ is difficult to maximize (e.g. due to lack of closed form FOC or lack of differentiability). Here, $d_k \in \mathbb{N}$ for each $j \in [k]$. Suppose that in each coordinate $j$, there exists a function $M_j(\boldsymbol{u}_j;\boldsymbol{v})$ that is easy to manipulate, such that (B1) $M_j(\boldsymbol{v}_j;\boldsymbol{v}) = O(\boldsymbol{v})$ and (B2) $M_j(\boldsymbol{u}_j;\boldsymbol{v}) \le O(\boldsymbol{w})$, where $\boldsymbol{w}^\top = \left(\boldsymbol{v}_1^\top, \dots, \boldsymbol{v}_{j-1}^\top, \boldsymbol{u}_j^\top, \boldsymbol{v}_{j+1}^\top, \dots, \boldsymbol{v}_k^\top\right)$, for all $\boldsymbol{v}^\top = \left(\boldsymbol{v}_1^\top, \dots, \boldsymbol{v}_k^\top\right) \in \mathbb{U}$. We say that $M_j(\boldsymbol{u}_j;\boldsymbol{v})$ is the $j$th blockwise minorizer of $O(\boldsymbol{v})$ at $\boldsymbol{v}$, or that $M_j(\boldsymbol{u}_j; \boldsymbol{v})$ minorizes $O(\boldsymbol{v})$ at $\boldsymbol{v}$ with respect to the $j$th coordinate.

Construct a blockwise-MM algorithm by firstly initializing it with some value $\boldsymbol{u}^{(0)}$. Next, at the $r$th iteration of the algorithm ($r \in \mathbb{N}$), set $\boldsymbol{u}_j^{(r)}$ to

$$\boldsymbol{u}_j^{(r)} = \begin{cases} \arg\max_{\boldsymbol{u}_j \in \mathbb{U}_j} \; M_j\left(\boldsymbol{u}_j; \boldsymbol{u}^{(r-1)}\right) & \text{if } j = (r \bmod k) + 1, \\ \boldsymbol{u}_j^{(r-1)} & \text{otherwise,} \end{cases} \tag{13}$$

for each $j \in [k]$, and then set $\boldsymbol{u}^{(r)\top} = \left(\boldsymbol{u}_1^{(r)\top}, \dots, \boldsymbol{u}_k^{(r)\top}\right)$. From (B1), (B2), and rule (13), we obtain

$$\begin{aligned} O\left(\boldsymbol{u}^{(r-1)}\right) &= M_{(r \bmod k)+1}\left(\boldsymbol{u}_{(r \bmod k)+1}^{(r-1)}; \boldsymbol{u}^{(r-1)}\right) \\ &\leq M_{(r \bmod k)+1}\left(\boldsymbol{u}_{(r \bmod k)+1}^{(r)}; \boldsymbol{u}^{(r-1)}\right) \\ &\leq O\left(\boldsymbol{u}^{(r)}\right) \end{aligned} \tag{14}$$

The sequence of inequalities (14) indicates that the sequence of blockwise-MM iterates $\left\{\boldsymbol{u}^{(r)}\right\}$ generates a sequence of objective evaluates $\left\{O\left(\boldsymbol{u}^{(r)}\right)\right\}$ that is monotonically increasing in $r$.

Denote directional derivative of $O(\boldsymbol{u})$ in the direction of $\boldsymbol{\delta}$ by

$$O'_{\boldsymbol{\delta}}(\boldsymbol{u}) = \liminf_{\lambda \downarrow 0} \frac{O(\boldsymbol{u} + \lambda\boldsymbol{\delta}) - f(\boldsymbol{u})}{\lambda},$$

and define a stationary point of $O(\boldsymbol{u})$ as any point $\boldsymbol{u}^*$ such that $O'_{\boldsymbol{\delta}}(\boldsymbol{u}^*) \geq 0$ for all $\boldsymbol{\delta}$ such that $\boldsymbol{u}^* + \boldsymbol{\delta} \in \mathbb{U}$. For all $j \in [k]$, make the following assumptions: (C1) $M_j(\boldsymbol{u}_j; \boldsymbol{u}) = O(\boldsymbol{u})$, for all $\boldsymbol{u} \in \mathbb{U}$; (C2) $M_j(\boldsymbol{u}_j; \boldsymbol{v}) \leq O(\boldsymbol{w})$, for all $\boldsymbol{u}_j \in \mathbb{U}_j$ and $\boldsymbol{v} \in \mathbb{U}$; (C3) $M'_{j,\boldsymbol{\delta}_j}(\boldsymbol{u}_j, \boldsymbol{v})\Big|_{\boldsymbol{u}_j = \boldsymbol{v}_j} = O'_{\boldsymbol{\delta}}(\boldsymbol{v})$, for all $\boldsymbol{\delta}^\top = \left(\boldsymbol{0}^\top, \dots, \boldsymbol{\delta}_j^\top, \dots, \boldsymbol{0}^\top\right)$ such that $\boldsymbol{u}_j + \boldsymbol{\delta}_j \in \mathbb{U}_j$; and $M_j(\boldsymbol{u}_j; \boldsymbol{v})$ is continuous in $\left(\boldsymbol{u}_j^\top, \boldsymbol{v}^\top\right)$. Assumptions (C1)–(C4) are validated if each $M_j(\boldsymbol{u}_j; \boldsymbol{v})$ is continuous and differentiable in $\left(\boldsymbol{u}_j^\top, \boldsymbol{v}^\top\right)$, and are blockwise minorizers that fulfill assumptions (B1) and (B2).

Say that a function $O(\boldsymbol{u})$ is regular at a point $\boldsymbol{v} \in \mathbb{U}$ if we have $O'_{\boldsymbol{\delta}}(\boldsymbol{v}) \geq 0$ for all $\boldsymbol{\delta}^\top = \left(\boldsymbol{\delta}_1^\top, \boldsymbol{\delta}_2^\top, \dots, \boldsymbol{\delta}_k^\top\right)$ with $O'_{\boldsymbol{\delta}_k^0}(\boldsymbol{v}) \geq 0$, where $\boldsymbol{\delta}_j^{0\top} = \left(\mathbf{0}^\top, \dots, \boldsymbol{\delta}_j^\top, \dots, \mathbf{0}^\top\right)$ and $\boldsymbol{v}_j + \boldsymbol{\delta}_j \in \mathbb{U}_j$, for all $j \in [k]$. Further say that $\boldsymbol{u}^*$ is a coordinate-wise maximum of $O(\boldsymbol{u})$ if $\boldsymbol{u}^* \in \mathbb{U}$ satisfies $O\left(\boldsymbol{u}^* + \boldsymbol{\delta}_j^0\right) \leq O(\boldsymbol{u}^*)$, for every $j \in [k]$ and $\boldsymbol{\delta}_j^0$ such that $\boldsymbol{u}^* + \boldsymbol{\delta}_j^0 \in \mathbb{U}$. Let $\boldsymbol{u}^{(\infty)} = \lim_{r\to\infty} \boldsymbol{u}^{(r)}$ be the limit point of a blockwise-MM algorithm defined via the update rule (13). The following theorem is available from Razaviyayn et al. (2013).

**Theorem 4.** *For all $j \in [k]$, assume that $M_j(\boldsymbol{u}_j; \boldsymbol{v})$ is quasi-concave in $\boldsymbol{u}_j \in \mathbb{U}_j$, for fixed $\boldsymbol{v} \in \mathbb{U}$, and fulfills Assumptions (C1)–(C4). Further assume that there is a unique solution to the problem:*

$$\arg \max_{\boldsymbol{u}_j \in \mathbb{U}_j} M_j(\boldsymbol{u}_j; \boldsymbol{v}),$$

*for each $j$, for any $\boldsymbol{v} \in \mathbb{U}$. If $\boldsymbol{u}^{(\infty)}$ is a limit point of a blockwise-MM algorithm defined by rule (13), then $\boldsymbol{u}^{(\infty)}$ is a coordinate-wise maximum of $O(\boldsymbol{u})$. Furthermore, if $\boldsymbol{u}^{(\infty)}$ is regular, then $\boldsymbol{u}^{(\infty)}$ is a stationary point of $O(\boldsymbol{u})$.*

Theorem 4 states that under some generous conditions, limit points of the blockwise-MM algorithm that are defined by rule (13) are stationary points of the objective. Furthermore, the convergence towards these stationary points is monotonically increasing in nature, with respect to the sequence of objective evaluations.

We furthermore note that global convergence results are also available for generalized blockwise-MM algorithms, where the iterates satisfy the relationships $M_j\left(\boldsymbol{u}^{(r)}; \boldsymbol{u}^{(r-1)}\right) \geq M_j\left(\boldsymbol{u}^{(r-1)}; \boldsymbol{u}^{(r-1)}\right)$ (for $j \in [k]$), but where $\boldsymbol{u}^{(r)}$ does not satisfy rule (13). In such cases, the algorithm retains its monotonicity with respect to the sequence of objective evaluations. It can further be shown that the limit points globally converge to a fixed-point via results such as that of Meyer (1976). However, the nature of the fixed-points cannot be stated in general and must be established on a case-to-case basis. As Theorem 4 is generally sufficient, we will not engage in further discussions of such results.

### 3.4 Example Blockwise-MM Algorithm

We consider the MQL estimation of an MoE model with soft-max gating functions and Gaussian regression experts from data $\{\boldsymbol{d}_i\}_{i=1}^n$, where $\boldsymbol{d}_i^\top = \left(\boldsymbol{x}_i^\top, y_i\right) \in \mathbb{R}^p \times \mathbb{R}$. Using characterization (3), the MoE model can be written as

$$\begin{aligned}\text{MoE}\left(y|\boldsymbol{x};\boldsymbol{\theta}\right) &= \sum_{z=1}^{g} \text{Gate}_z\left(\boldsymbol{x};\boldsymbol{\gamma}\right)\text{Expert}_z\left(y|\boldsymbol{x};\boldsymbol{\eta}_z\right)\\ &= \sum_{z=1}^{g} \frac{\exp\left(\alpha_{z0} + \boldsymbol{\alpha}_z^\top \boldsymbol{x}\right)}{\sum_{\zeta=1}^{g}\exp\left(\alpha_{\zeta 0} + \boldsymbol{\alpha}_\zeta^\top \boldsymbol{x}\right)}\phi\left(\boldsymbol{y};\beta_{0z} + \boldsymbol{\beta}_z^\top \boldsymbol{x}, \sigma_z^2\right),\end{aligned}$$

where $\beta_{0z} \in \mathbb{R}$, $\boldsymbol{\beta}_z \in \mathbb{R}^p$, $\sigma_z^2 > 0$, and $\boldsymbol{\eta}_z^\top = \left(\beta_{z0}, \boldsymbol{\beta}_z^\top, \sigma_z^2\right)$, for each $z \in [g]$. The log-quasi-likelihood function can then be written as

$$Q_n\left(\boldsymbol{\theta}\right) = \sum_{i=1}^{n}\log\sum_{z=1}^{g}\frac{\exp\left(\alpha_{z0} + \boldsymbol{\alpha}_z^\top \boldsymbol{x}_i\right)}{\sum_{\zeta=1}^{g}\exp\left(\alpha_{\zeta 0} + \boldsymbol{\alpha}_\zeta^\top \boldsymbol{x}_i\right)}\phi\left(y_i;\beta_{0z} + \boldsymbol{\beta}_z^\top \boldsymbol{x}_i, \sigma_z^2\right). \tag{15}$$

We observe that although (15) is smooth in all coordinates of $\boldsymbol{\theta}$, it is in the log-sum-exp form and thus a closed form solution to the usual FOC cannot be obtained. We thus turn to constructing a blockwise-MM algorithm for obtaining suitable roots of (15).

Let $\boldsymbol{u}$ and $\boldsymbol{v}$ be such that $u_j > 0$ and $v_j > 0$ for each $j \in [k]$. We can minorize $O\left(\boldsymbol{u}\right) = \log\sum_{j=1}^{k} u_j$ in all coordinates, simultaneous, by the minorizer

$$M\left(\boldsymbol{u};\boldsymbol{v}\right) = \sum_{j=1}^{k}\frac{v_j}{\sum_{z=1}^{k} v_z}\log u_j - \sum_{j=1}^{k}\frac{v_j}{\sum_{z=1}^{k} v_z}\log\frac{v_j}{\sum_{z=1}^{k} v_z}, \tag{16}$$

from Zhou & Lange (2010). Applying (16), we obtain the minorizer

$$\begin{aligned} R\left(\boldsymbol{\theta};\boldsymbol{\theta}^{(r-1)}\right) &= \sum_{i=1}^{n}\sum_{z=1}^{g}\tau_z\left(\boldsymbol{d}_i;\boldsymbol{\theta}^{(r-1)}\right)\left(\alpha_{z0}+\boldsymbol{\alpha}_z^\top\boldsymbol{x}_i\right) \\ &-\sum_{i=1}^{n}\log\sum_{\zeta=1}^{g}\exp\left(\alpha_{\zeta 0}+\boldsymbol{\alpha}_\zeta^\top\boldsymbol{x}_i\right) \\ &-\frac{1}{2}\sum_{i=1}^{n}\sum_{z=1}^{g}\tau_z\left(\boldsymbol{d}_i;\boldsymbol{\theta}^{(r-1)}\right)\log\sigma_z^2 \\ &-\frac{1}{2}\sum_{i=1}^{n}\sum_{z=1}^{g}\tau_z\left(\boldsymbol{d}_i;\boldsymbol{\theta}^{(r-1)}\right)\frac{\left(y-\beta_{0z}-\boldsymbol{\beta}_z^\top\boldsymbol{x}_i\right)^2}{\sigma_z^2} \end{aligned} \tag{17}$$

at $\boldsymbol{\theta}^{(r-1)}$ (in all coordinates), where

$$\tau_z\left(\boldsymbol{d}_i;\boldsymbol{\theta}\right)=\frac{\mathrm{Gate}_z\left(\boldsymbol{x}_i;\boldsymbol{\gamma}\right)\mathrm{Expert}_z\left(y_i|\boldsymbol{x}_i;\boldsymbol{\eta}_z\right)}{\mathrm{MoE}\left(y_i|\boldsymbol{x}_i;\boldsymbol{\theta}\right)}$$

and

$$C\left(\boldsymbol{\theta}^{(r-1)}\right)=-\frac{n}{2}\log 2\pi-\sum_{i=1}^{n}\sum_{z=1}^{g}\tau_z\left(\boldsymbol{d}_i;\boldsymbol{\theta}^{(r-1)}\right)\log\tau_z\left(\boldsymbol{d}_i;\boldsymbol{\theta}^{(r-1)}\right)$$

is a constant that does not depend on $\boldsymbol{\theta}$.

Define $\tilde{\boldsymbol{x}}_i^\top=\left(1,\boldsymbol{x}_i^\top\right)$, for each $i\in[n]$, and $\tilde{\boldsymbol{\alpha}}_z^\top=\left(\alpha_{0z},\boldsymbol{\alpha}_z^\top\right)$ and $\tilde{\boldsymbol{\beta}}_z^\top=\left(\beta_{0z},\boldsymbol{\beta}_z^\top\right)$, for each $z\in[g]$. We can rewrite (17) as

$$R\left(\boldsymbol{\theta};\boldsymbol{\theta}^{(r-1)}\right)=R_0\left(\boldsymbol{\theta};\boldsymbol{\theta}^{(r-1)}\right)+R_g\left(\boldsymbol{\theta};\boldsymbol{\theta}^{(r-1)}\right)+C\left(\boldsymbol{\theta}^{(r-1)}\right),$$

where

$$\begin{aligned} R_0\left(\boldsymbol{\theta};\boldsymbol{\theta}^{(r-1)}\right) &= \sum_{i=1}^{n}\sum_{z=1}^{g}\tau_z\left(\boldsymbol{d}_i;\boldsymbol{\theta}^{(r-1)}\right)\tilde{\boldsymbol{\alpha}}_z^\top\tilde{\boldsymbol{x}}_i \\ &-\sum_{i=1}^{n}\log\sum_{\zeta=1}^{g}\exp\left(\tilde{\boldsymbol{\alpha}}_\zeta^\top\tilde{\boldsymbol{x}}_i\right) \end{aligned}$$

and

$$R_g\left(\boldsymbol{\theta};\boldsymbol{\theta}^{(r-1)}\right) = -\frac{1}{2}\sum_{i=1}^{n}\sum_{z=1}^{g}\tau_z\left(\boldsymbol{d}_i;\boldsymbol{\theta}^{(r-1)}\right)\log\sigma_z^2$$
$$-\frac{1}{2}\sum_{i=1}^{n}\sum_{z=1}^{g}\tau_z\left(\boldsymbol{d}_i;\boldsymbol{\theta}^{(r-1)}\right)\frac{\left(y-\tilde{\boldsymbol{\beta}}_z^\top\boldsymbol{x}_i\right)^2}{\sigma_z^2}.$$

Let $\boldsymbol{\theta}^\top = \left(\boldsymbol{\theta}_1^\top,\dots,\boldsymbol{\theta}_g^\top\right)$ be a partitioning of the coordinates of $\boldsymbol{\theta}$, where $\boldsymbol{\theta}_z = \tilde{\boldsymbol{\alpha}}_z^\top$, for $z\in[g-1]$, and $\boldsymbol{\theta}_g^\top = \left(\tilde{\boldsymbol{\beta}}_1^\top,\dots,\tilde{\boldsymbol{\beta}}_g^\top,\sigma_1^2,\dots,\sigma_g^2\right)$. For twice differentiable and concave functions $O(\boldsymbol{u})$, Bohning & Lindsay (1988) proposed the minorizer at $\boldsymbol{v}$ (in all coordinates)

$$M(\boldsymbol{u};\boldsymbol{v}) = O(\boldsymbol{v}) + (\boldsymbol{u}-\boldsymbol{v})^\top\nabla O(\boldsymbol{v}) + \frac{1}{2}(\boldsymbol{u}-\boldsymbol{v})^\top\mathbf{H}(\boldsymbol{u}-\boldsymbol{v}), \tag{18}$$

where $\mathbf{H}-\mathrm{Hess}(O)(\boldsymbol{u})$ is negative semi-definite for all $\boldsymbol{u}\in\mathbb{U}$ and $\mathbf{H}$ is positive definite. Here, $\nabla(\cdot)$ is the gradient operator and $\mathrm{Hess}(\cdot)$ is the Hessian operator. Let

$$R_z\left(\boldsymbol{\theta}_z;\boldsymbol{\theta}^{(r)}\right) = R_0\left(\boldsymbol{\vartheta}_z^{(r-1)};\boldsymbol{\theta}^{(r-1)}\right),$$

where $\boldsymbol{\vartheta}^{(r-1)\top} = \left(\boldsymbol{\theta}_1^{(r-1)\top},\dots,\boldsymbol{\theta}_{z-1}^{(r-1)\top},\boldsymbol{\theta}_z^\top,\boldsymbol{\theta}_{z+1}^{(r-1)\top},\dots,\boldsymbol{\theta}_{g-1}^{(r-1)\top},\boldsymbol{\theta}_g^{(r-1)\top}\right)$. Applying (18) to $R_z\left(\boldsymbol{\theta}_z;\boldsymbol{\theta}^{(r-1)}\right)$, we obtain the coordinate-wise minorizer of (15):

$$\begin{aligned}S_z\left(\boldsymbol{\theta}_z;\boldsymbol{\theta}^{(r-1)}\right) &= R_0\left(\boldsymbol{\theta}^{(r-1)};\boldsymbol{\theta}^{(r-1)}\right) + \nabla R_z\left(\boldsymbol{\theta}_z;\boldsymbol{\theta}^{(r-1)}\right)\Big|_{\boldsymbol{\theta}_z=\boldsymbol{\theta}_z^{(r-1)}}\\ &\quad -\frac{1}{8}\left(\boldsymbol{\theta}_z-\boldsymbol{\theta}_z^{(r-1)}\right)^\top\mathbf{H}\left(\boldsymbol{\theta}_z-\boldsymbol{\theta}_z^{(r-1)}\right)\\ &\quad + R_g\left(\boldsymbol{\theta}^{(r-1)};\boldsymbol{\theta}^{(r-1)}\right) + C\left(\boldsymbol{\theta}^{(r-1)}\right)\end{aligned} \tag{19}$$

for each $z\in[g-1]$, where $\mathbf{H}=\sum_{i=1}^n\tilde{\boldsymbol{x}}_i\tilde{\boldsymbol{x}}_i^\top$ and

$$\nabla R_z\left(\boldsymbol{\theta}_z;\boldsymbol{\theta}^{(r-1)}\right) = \sum_{i=1}^{n}\left[\tau_z\left(\boldsymbol{d}_i;\boldsymbol{\theta}^{(r-1)}\right) - \mathrm{Gate}_z\left(\boldsymbol{x}_i;\tilde{\boldsymbol{\gamma}}_z^{(r-1)}\right)\right]\boldsymbol{x}_i.$$

This is obtained by noting that

$$\text{Hess}\ \left(R_z\right)\left(\boldsymbol{\theta}_z;\boldsymbol{\theta}^{(r-1)}\right)=-\sum_{i=1}^{n}\text{Gate}_z\left(\boldsymbol{x}_i;\tilde{\boldsymbol{\gamma}}_z^{(r-1)}\right)\left[1-\text{Gate}_z\left(\boldsymbol{x}_i;\tilde{\boldsymbol{\gamma}}_z^{(r-1)}\right)\right]\tilde{\boldsymbol{x}}_i\tilde{\boldsymbol{x}}_i^{\top},$$

where $\tilde{\boldsymbol{\gamma}}_z^{(r-1)\top}=\left(\boldsymbol{\theta}_1^{(r-1)\top},\ldots,\boldsymbol{\theta}_{z-1}^{(r-1)\top},\boldsymbol{\theta}_z^{\top},\boldsymbol{\theta}_{z+1}^{(r-1)\top},\ldots,\boldsymbol{\theta}_{g-1}^{(r-1)\top}\right)$, and that $a\left(1-a\right)\leq 1/4$ for any $a\in(0,1)$.

Notice that (19) is a quadratic and thus is concave and has a unique maximizer with respect to $\boldsymbol{\theta}_z$. We can obtain the maximizer by solving the FOC $\nabla S_z\left(\boldsymbol{\theta}_z;\boldsymbol{\theta}^{(r-1)}\right)=\mathbf{0}$, which yields the solution

$$\boldsymbol{\theta}_z^{(r)}=4\times\mathbf{H}^{-1}\ \nabla R_z\left(\boldsymbol{\theta}_z;\boldsymbol{\theta}^{(r-1)}\right)\Big|_{\boldsymbol{\theta}_z=\boldsymbol{\theta}_z^{(r-1)}}+\boldsymbol{\theta}_z^{(r-1)},\tag{20}$$

to the problem

$$\arg\max_{\boldsymbol{\theta}_z}\ S_z\left(\boldsymbol{\theta}_z;\boldsymbol{\theta}^{(r-1)}\right),$$

for each $z\in[g-1]$. Recall that $\tilde{\boldsymbol{\alpha}}_g=\mathbf{0}$ and thus does not require updating.

Next, a minorizer of (15) in $\boldsymbol{\theta}_g$ can be obtained by simply holding all other coordinates constant. That is,

$$S_g\left(\boldsymbol{\theta}_g;\boldsymbol{\theta}^{(r-1)}\right)=R_0\left(\boldsymbol{\theta}^{(r-1)};\boldsymbol{\theta}^{(r-1)}\right)+R_g\left(\boldsymbol{\vartheta}_g^{(r-1)};\boldsymbol{\theta}^{(r-1)}\right)+C\left(\boldsymbol{\theta}^{(r-1)}\right)$$

is a coordinate-wise minorizer of (15) in $\boldsymbol{\theta}_g$. The quasi-concavity of $S_g\left(\boldsymbol{\theta}_g;\boldsymbol{\theta}^{(r-1)}\right)$ and the solution to its FOC $\nabla S_g\left(\boldsymbol{\theta}_g;\boldsymbol{\theta}^{(r-1)}\right)=\mathbf{0}$ can be obtained via slight modifications to the results of Nguyen & McLachlan (2015). For completeness, the solution $\boldsymbol{\theta}_z^{(r)}$ containing

$$\tilde{\boldsymbol{\beta}}_z^{(r)}=\left[\sum_{i=1}^{n}\tau_z\left(\boldsymbol{d}_i;\boldsymbol{\theta}^{(r-1)}\right)\tilde{\boldsymbol{x}}_i\tilde{\boldsymbol{x}}_i^{\top}\right]^{-1}\sum_{i=1}^{n}\tau_z\left(\boldsymbol{d}_i;\boldsymbol{\theta}^{(r-1)}\right)y_i\boldsymbol{x}_i,\tag{21}$$

and

$$\sigma_z^{2(r)} = \frac{\sum_{i=1}^{n} \tau_z\left(\boldsymbol{d}_i;\boldsymbol{\theta}^{(r-1)}\right)\left(y_i - \tilde{\boldsymbol{\beta}}_z^{(r)\top}\boldsymbol{x}_i\right)^2}{\sum_{i=1}^{n} \tau_z\left(\boldsymbol{d}_i;\boldsymbol{\theta}^{(r-1)}\right)}, \quad (22)$$

for each $z \in [g]$, uniquely solves the problem

$$\arg\max_{\boldsymbol{\theta}_g} \; S_g\left(\boldsymbol{\theta}_z;\boldsymbol{\theta}^{(r-1)}\right).$$

Together updates (20)–(22) can be applied within rule (13) in order to generate an blockwise-MM algorithm for obtaining the MQL estimator of (15). Since all blockwise solutions are unique and each blockwise minorizers is quasiconcave, we obtain the full conclusion of Theorem 4, as the objective function is smooth and thus regularity is not an issue.

For each initialization $\boldsymbol{\theta}^{(0)}$, the blockwise-MM algorithm tends towards a single solution. Unfortunately, like many mixture-type models, the log-quasi-likelihood of MoE models tend to be highly multimodal. As such, numerous initializations should be considered in order to locate a good local maximum, which can then be considered as candidates for the MQL estimator. One technique for choosing good initializations is that of McLachlan (1988).

We note that although we have derived an algorithm that is entirely within the MM framework, it is possible to replace some of the updates with numerical or alternative optimization schemes that are outside of the MM paradigm. For example, in Ng & McLachlan (2004), an Newton procedure was utilized to update $\boldsymbol{\theta}_z$ for each $z \in [g-1]$, upon firstly minorizing (15) by (16). As long as the alternative schemes yields solutions that satisfy rule (13) for some notion of blockwise minorization functions, the resulting hybrid blockwise-MM algorithms that are produced will retain the desirable properties bestowed by Theorem 4.

### 3.5 Asymptotic Properties of the Maximum Quasi-Likelihood Estimator

We now consider the asymptotic properties of the MQL estimator. The consistency and asymptotic normality of the MQL estimator for the MoE model with Gaussian experts was proved in Zeevi et al. (1998). Further results for GLM experts were obtained in Jiang & Tanner (2000). We shall provide a general scheme for deriving such results for arbitrary MoE models, using the extremum estimation concept of Amemiya (1985).

Let $O_n(\boldsymbol{u}) = O(\boldsymbol{u}; \boldsymbol{D}_1, \ldots, \boldsymbol{D}_n)$ be an arbitrary objective function that takes random inputs $\{\boldsymbol{D}_i\}_{i=1}^n$ and is parameterized by $\boldsymbol{\theta}$. Suppose that we wish to obtain the properties of the extremum estimator

$$\hat{\boldsymbol{u}}_n = \arg\max_{\boldsymbol{u} \in \mathbb{U}} O_n(\boldsymbol{u}), \tag{23}$$

for some Euclidean subset $\mathbb{U} \subset \mathbb{R}^d$ ($d \in \mathbb{N}$). Suppose that there is some $\boldsymbol{u}_0$ that naturally connects $O_n(\boldsymbol{u})$ to the DGP of $\{\boldsymbol{D}_i\}_{i=1}^n$. We say that $\hat{\boldsymbol{u}}_n$ is consistent if it converges to $\boldsymbol{u}_0$ in probability. The following theorem of Amemiya (1985) provides a simple set of assumptions that can be used to establish the consistency of (23).

**Theorem 5.** *Make the following assumptions: (D1) let $\mathbb{U}$ be open; (D2) let $O_n(\boldsymbol{u})$ be measurable in $\{\boldsymbol{D}_i\}_{i=1}^n$ for all $\boldsymbol{u} \in \mathbb{U}$, and let $\nabla O_n(\boldsymbol{u})$ exist and be continuous in an open neighborhood of $\boldsymbol{u}_0 \in \mathbb{U}$; and (D3) let $n^{-1}O_n(\boldsymbol{u})$ converge to a non-stochastic function $O(\boldsymbol{u})$ in probability uniformly in $\boldsymbol{u}$, in an open neighborhood of $\boldsymbol{u}_0$, and let let $O(\boldsymbol{u})$ attain a strict local maximum at a root $\boldsymbol{u}_0$. If (D1)–(D3) are fulfilled and*

$$\mathbb{U}_n = \{\boldsymbol{u} : \nabla O_n(\boldsymbol{u}) = \boldsymbol{0} \text{ and } \boldsymbol{u} \text{ is a strict local maximum}\}, \tag{24}$$

*then*

$$\lim_{n\to\infty} \mathbb{P}\left(\inf_{\boldsymbol{u}\in\mathbb{U}_n} \|\boldsymbol{u}-\boldsymbol{u}_0\| > \epsilon\right) = 0,$$

*for any $\epsilon > 0$. Here, $\mathbb{U}_n = \{$arbitrary element of $\mathbb{U}\}$ when definition (24) results in the empty set.*

The conclusion of Theorem 5 is that there exists a consistent root, $\boldsymbol{u}_0$, that is a local maximizer of the objective function $O_n(\boldsymbol{u})$. The result is useful in MoE modeling due to the general lack of universal identifiability of MoE models (cf. Jiang & Tanner), which leads to multiple roots corresponding to the same underlying DGP. Furthermore, the result is also useful due to the highly multimodal nature of MoE log-quasi-likelihood functions. We note that in general, it is not obvious which of many roots is the consistent one that leads to the best approximation of the underlying DGP. Following a suggestion of Amemiya (1985), we can gain some confidence regarding a particular root based on its reasonability from a scientific or contextual perspective, or if it is the limit point of an algorithm when initialized from numerous starting points. A theoretical method for choosing between multiple roots in MQL estimation was proposed by Gan & Jiang (1999). Upon establishing the consistency of a root $\hat{\boldsymbol{u}}_n$, we can then deduce its asymptotic normality via the following theorem of Amemiya (1985).

**Theorem 6.** *Make the following assumptions: (E1) let $Hess(O_n)(\boldsymbol{u})$ exist and be continuous in a convex neighborhood of $\boldsymbol{u}_0$; (E2) let $n^{-1}Hess(O_n)(\boldsymbol{u}_n^*)$ converge to a finite and non-singular matrix*

$$\boldsymbol{I}_1(\boldsymbol{u}_0) = \lim_{n\to\infty} \mathbb{E}n^{-1}Hess(O_n)(\boldsymbol{u}_0)$$

*in probability for any sequence $\boldsymbol{u}_n^*$ that converges to $\boldsymbol{u}_0$ in probability; (E3) let $n^{-1/2}\left.\nabla O_n(\boldsymbol{u})\right|_{\boldsymbol{u}=\boldsymbol{u}_0}$ be asymptotically normal with mean $\mathbf{0}$ and covariance* *matrix* $\boldsymbol{I}_2(\boldsymbol{u}_0)$, *where*

$$\boldsymbol{I}_2(\boldsymbol{u}_0) = \lim_{n\to\infty} \mathbb{E} n^{-1} \left.\nabla O_n(\boldsymbol{u})\right|_{\boldsymbol{u}=\boldsymbol{u}_0} \left.\nabla^\top O_n(\boldsymbol{u})\right|_{\boldsymbol{u}=\boldsymbol{u}_0}.$$

*If* $\{\hat{\boldsymbol{u}}_n\}_{n=1}^{\infty}$ *is a sequence that is obtained by choosing an element from* $\mathbb{U}_n$*, for each* $n \in \mathbb{N}$*, such that* $\hat{\boldsymbol{u}}_n$ *converges to* $\boldsymbol{u}_0$ *in probability, then* $n^{-1/2}(\hat{\boldsymbol{u}}_n - \boldsymbol{u}_0)$ *is asymptotically normal with mean* $\mathbf{0}$ *and covariance matrix*

$$\boldsymbol{I}(\boldsymbol{u}_0) = \boldsymbol{I}_1^{-1}(\boldsymbol{u}_0)\, \boldsymbol{I}_2(\boldsymbol{u}_0)\, \boldsymbol{I}_1^{-1}(\boldsymbol{u}_0).$$

Theorem 6 allows for the construction of asymptotic hypothesis tests and confidence intervals regarding the obtained consistent root $\hat{\boldsymbol{u}}_n$. Such tests and intervals can be constructed via results such as those from Hayashi (2000, Sec. 7.4). For the purpose of hypothesis testing, knowledge of $\boldsymbol{u}_0$ is assumed. However, when constructing confidence intervals, the DGP is generally unknown. Thus, one must estimate $\boldsymbol{I}(\boldsymbol{u}_0)$ in such constructions. When $O_n(\boldsymbol{u}) = \sum_{i=1}^n o(\boldsymbol{D}_i; \boldsymbol{u})$, a natural estimator for $\boldsymbol{I}(\boldsymbol{u}_0)$ is

$$\hat{\boldsymbol{I}}_n(\hat{\boldsymbol{u}}_n) = \hat{\boldsymbol{I}}_{1,n}^{-1}(\hat{\boldsymbol{u}}_n)\, \hat{\boldsymbol{I}}_{2,n}(\hat{\boldsymbol{u}}_n)\, \hat{\boldsymbol{I}}_{1,n}^{-1}(\hat{\boldsymbol{u}}_n), \tag{25}$$

where

$$\hat{\boldsymbol{I}}_{1,n}(\hat{\boldsymbol{u}}_n) = n^{-1} \sum_{i=1}^{n} \mathrm{Hess}(o)(\boldsymbol{D}_i; \hat{\boldsymbol{u}}_n)$$

and

$$\hat{\boldsymbol{I}}_{2,n}(\hat{\boldsymbol{u}}_n) = n^{-1} \sum_{i=1}^{n} \left.\nabla o_n(\boldsymbol{D}_i; \boldsymbol{u})\right|_{\boldsymbol{u}=\hat{\boldsymbol{u}}_n} \left.\nabla^\top o_n(\boldsymbol{D}_i; \boldsymbol{u})\right|_{\boldsymbol{u}=\hat{\boldsymbol{u}}_n}.$$

Results such as the one of Boos & Stefanski (2013, Thm. 7.3) can be used to establish the validity of (25).

Fix $g \in \mathbb{N}$ and let $\mathbb{U}$ be the space of valid values that $\boldsymbol{\theta}$ can take in the log-quasi-likelihood function 15. Set $O_n(\boldsymbol{\theta}) = Q_n(\boldsymbol{\theta})$ and let $Q_n(\boldsymbol{\theta}) = \sum_{i=1}^{n} q(\boldsymbol{\theta}; \boldsymbol{D}_i)$, where $q(\boldsymbol{\theta}; \boldsymbol{d}_i) = \log \text{MoE}(\boldsymbol{y}_i | \boldsymbol{x}_i; \boldsymbol{\theta})$ with gating function and experts as per Section 3.4 (the MoE model has gating functions of form (4) and the experts are of form (6)). For convenience, suppose that $\{\boldsymbol{D}_i\}_{i=1}^{n}$ is an IID sample from a DGP with continuous PDF over a compact support. By definition of the MoE model and its parameter vector $\boldsymbol{\theta}$, the space of valid values $\mathbb{U}$ is an open subset of a Euclidean space and hence validates assumption (D1). Since the MoE is constructed from gating functions of form (4) and experts of form (6), it is continuously differentiable, and thus its logarithm is also continuously differentiable. Since the PDF of $\boldsymbol{D}_i$ is continuous and the support is compact, $O_n(\boldsymbol{\theta})$ is also measurable, thus validating (D2). Since the PDF of $\boldsymbol{D}_i$ is continuous and the support is compact, and since $\{\boldsymbol{D}_i\}_{i=1}^{n}$ is an IID sample, it is procedural to validate (D3) for $O(\boldsymbol{\theta}) = \mathbb{E}q(\boldsymbol{D}_i; \boldsymbol{\theta})$ via a uniform law of large numbers such as that of Jennrich (1969). We therefore obtain the conclusion of Theorem 5 for the MQL estimator $\hat{\boldsymbol{\theta}}_n$ of the MoE model from Section 3.4.

Further, we note that the MoE model above also has continuous Hessian for all valid inputs from $\mathbb{U}$ and thus (E1) is valid. Assumption (E3) is valid because (D3) implies that $\boldsymbol{u}_0$ solves $\nabla \mathbb{E}q(\boldsymbol{D}_i; \boldsymbol{\theta}) = \mathbf{0}$ and because we can swap the gradient and expectation operator since the PDF of $\boldsymbol{D}_i$ is continuous and the support is compact. An application of the multivariate central limit theorem yields the desired result. Finally, we must make assumption that $\mathbb{E}\text{Hess}(q)(\boldsymbol{D}_i; \boldsymbol{\theta}_0)$ is non-singular in order to validate (E2). Such an assumption is standard in the literature and cannot be done away with in general. We therefore have the conclusion of Theorem 6 for the MQL estimator $\hat{\boldsymbol{\theta}}_n$ of the MoE model from Section 3.4.

## 3.6 Choice of Number of Components

Thus far, we have assumed that the number of experts (components) $g \in \mathbb{N}$ is some known constant. However, in reality, its value in the best MoE approximation of the DGP is as unknown as is the value of the parameter vector $\boldsymbol{\theta}$. In the literature, a popular method for choosing the value of $g$ among many candidates is to use an information criterion such as Akaike information criterion (AIC; Akaike, 1974), the Bayesian information criterion (BIC; Schwarz, 1978), or the Integrated complete-likelihood information criterion (ICL; Biernacki et al., 2000).

The aforementioned criteria have been implemented in articles such as Grun & Leisch (2007), Grun & Leisch (2008), Chamroukhi et al. (2009), and Nguyen & McLachlan (2016). It is notable that until recently, the only theoretical justification for any of these criteria is for the BIC, which was demonstrated to be consistent in Olteanu & Rynkiewicz (2011). The MoE model results from Olteanu & Rynkiewicz (2011) can be viewed as extensions of the marginal mixture model results from Keribin (2000). Unfortunately, the result of Olteanu & Rynkiewicz (2011) is difficult to state concisely. A more recent approach by Baudry (2015) allows for a much simpler statement of an information criterion consistency theorem. Using the notation from Section 3.5, we paraphrase Baudry (2015, Thm. 8.1) below.

**Theorem 7.** *Let* $\{\mathbb{U}_g\}_{g=1}^{G}$ *be a set of parameter spaces, for any* $g \in [G]$*, such that* $\mathbb{U}_g \subset \mathbb{R}^{d_g}$*, where* $d_g \in \mathbb{N}$ *for each* $g$*, and* $d_1 \leq \cdots \leq d_G$*. Further let*

$$\mathbb{U}_0^{[g]} = \left\{ \boldsymbol{u} : \boldsymbol{u} = \arg\max_{\boldsymbol{u} \in \mathbb{U}_g} \mathbb{E}O\left(\boldsymbol{u}\right) \right\}$$

*for some* $O_n\left(\boldsymbol{u}\right) = O\left(\boldsymbol{u}; \boldsymbol{D}_1, \ldots, \boldsymbol{D}_n\right)$ *that is a function of both* $\boldsymbol{u} \in \mathbb{U}_g$ *and*

$\{\boldsymbol{D}_i\}_{i=1}^n$. *Make the following assumptions: (F1) let*

$$\mathbb{G}_0 = \left\{ g : g = \arg\max_{g \in [G]} \mathbb{E} O\left(\boldsymbol{u}_0^{[g]}\right), \, \boldsymbol{u}_0^{[g]} \in \mathbb{U}_0^{[g]} \right\}$$

*and assume that* $g = \min \mathbb{G}_0$*; (F2) for all* $g \in [G]$, $\hat{\boldsymbol{u}}_n^{[g]} \in \mathbb{U}_{[n]}^{[g]}$, *where*

$$\mathbb{U}_{[n]}^{[g]} = \left\{ \boldsymbol{u} : O_n(\boldsymbol{u}) \geq O_n\left(\boldsymbol{u}_0^{[g]}\right), \, O_n(\boldsymbol{u}) \to \mathbb{E} O\left(\boldsymbol{u}_0^{[g]}\right) \text{ in probability} \right\};$$

*(F3) for all* $g \in [g]$*, define* $pen_n(g)$ *to be such that* $pen_n(g) > 0$, $n^{-1} pen_n(g) \to 0$ *in probability, as* $n \to \infty$*, and*

$$pen_n(g) - pen_n(g^*) \to \infty$$

*in probability, as* $n \to \infty$*, when* $g > g^*$*; (F4)*

$$O_n\left(\hat{\boldsymbol{u}}_n^{[g_0]}\right) - O_n\left(\hat{\boldsymbol{u}}_n^{[g]}\right) \to C$$

*in probability, where* $C$ *is a constant, for any* $g \in \mathbb{G}_0$*. If (F1)–(F4) are fulfilled and if selection of* $g$ *is based upon the generic information criterion:*

$$\hat{g}_n = \min \left\{ g : g = \arg\max_{g \in [G]} \left[ O_n\left(\hat{\boldsymbol{u}}_n^{[g]}\right) - pen_n(g) \right] \right\}, \tag{26}$$

*then* $\mathbb{P}(\hat{g}_n \neq g_0) \to 0$ *as* $n \to \infty$*.*

Twice the negative of $O_n\left(\boldsymbol{u}_n^{[g]}\right) - \text{pen}_n(g)$, in (26), is often referred to as the information criterion. Although tedious, the assumptions of Theorem 7 are generally valid. Assumption (F1) states that we are searching for a parsimonious model, and (F2) is valid if the hypothesis of Theorem 5 are valid for each $g \in [G]$. Assumption (F3) states that the constructed information criterion must involve a penalty (at the discretion of the investigator) that becomes smaller as more observations are observed and that is capable of ordering different complexities of models. Assumption (F4) is difficult to rationalize, although it can be validated by application of Baudry (2015, Cor. 8.2).

Consider the penalty

$$\text{pen}_n(g) = \dim(\boldsymbol{\theta}) \log n, \tag{27}$$

where $\dim(\cdot)$ computes the dimension (i.e. number of elements) of a vector. This is the BIC penalty function. For the soft-max gated mixture of Gaussian regression experts, we can show that

$$\dim(\boldsymbol{\theta}) = (3 + 2p) g - p - 1.$$

We can validate (F3) by noting that $\lim_{n\to\infty} n^{-1} \log n = 0$ and that $\log n$ is strictly increasing. Using Baudry (2015, Cor. 8.2) to validate (F4), and the assumptions made in the example from Section 3.5 to validate (F2), we can show that the BIC (i.e. rule (26) with penalty (27)) consistently selects the most parsimonious MoE model with gating functions of form (4) and experts of form (6), with respect to number of components $g$.

# 4 Applications of Mixture-of-Experts Models

## 4.1 Classification

We can conduct classification via MoE modeling by using an MoE model with multinomial logistic experts. That is, suppose that we observe data $\{\boldsymbol{D}_i\}_{i=1}^n$, where $\boldsymbol{D}_i^\top = (\boldsymbol{X}_i^\top, Y_i) \in \mathbb{X} \times [K]$, for some $K \in \mathbb{N}$. Conditioned on $\boldsymbol{X}_i = \boldsymbol{x}_i$, suppose that the PMF of $Y_i$ can be best approximated by an MoE of form (3) with multinomial logistic expert functions of form

$$\text{Expert}_z\left(y|\boldsymbol{x},\boldsymbol{\eta}_z\right)=\prod_{l=1}^{K}\left[\frac{\exp\left(\beta_{zl0}+\boldsymbol{\beta}_{zl}^{\top}\boldsymbol{x}\right)}{\sum_{\ell=1}^{K}\exp\left(\beta_{z\ell 0}+\boldsymbol{\beta}_{z\ell}^{\top}\boldsymbol{x}\right)}\right]^{\mathbb{I}(y=l)}, \tag{28}$$

where $\beta_{zl0}\in\mathbb{R}$ and $\boldsymbol{\beta}_{zl}\in\mathbb{R}^p$ for each $z\in[g]$ and $l\in[K-1]$, and $\beta_{zK0}$ and $\boldsymbol{\beta}_{zK}=\mathbf{0}$ for each $z\in[g]$. We set

$$\boldsymbol{\eta}_z^{\top}=\left(\beta_{z10},\boldsymbol{\beta}_{z1}^{\top},\dots,\beta_{z,K-1,0},\boldsymbol{\beta}_{z,K-1}^{\top}\right)$$

for each $z$. Here, $\mathbb{I}(A)$ is the indicator function that takes value 1 if proposition $A$ is true and 0 otherwise.

Respectively, let $g_0$ and $\boldsymbol{\theta}_0$ be the number of components and parameter vector that best approximates the DGP of interest. The MoE model of form (3), with experts of form (28), has the probabilistic interpretation

$$\mathbb{P}\left(Y=y|\boldsymbol{X}=\boldsymbol{x}\right)=\text{MoE}\left(y|\boldsymbol{x};\boldsymbol{\theta}_0\right). \tag{29}$$

Let $\boldsymbol{d}^{\top}=\left(\boldsymbol{x}^{\top},y\right)$ be an arbitrary data point that is generated via the same DGP as that of interest, and suppose that we only have knowledge of $\boldsymbol{x}$ and wish to estimate $y$. Using interpretation (29), we can obtain the MAP (maximum a posteriori probability) rule for classification:

$$\begin{aligned}\hat{y}&=\arg\max_{y\in[K]}\ \text{MoE}\left(y|\boldsymbol{x};\boldsymbol{\theta}_0\right)\\&=\arg\max_{y\in[K]}\sum_{z=1}^{g_0}\text{Gate}_z\left(\boldsymbol{x};\boldsymbol{\gamma}_0\right)\text{Expert}_z\left(y|\boldsymbol{x};\boldsymbol{\eta}_{0,z}\right).\end{aligned}$$

If $g_0$ and $\boldsymbol{\theta}_0$ are unknown, then we can estimate these quantities by $\hat{g}_n$ and $\hat{\boldsymbol{\theta}}_n$,

respectively, in order to obtain the plugin-MAP rule

$$\begin{aligned}\hat{y} &= \arg\max_{y\in[K]} \operatorname{MoE}\left(y|\boldsymbol{x};\hat{\boldsymbol{\theta}}_n\right) \\ &= \arg\max_{y\in[K]} \sum_{z=1}^{\hat{g}_n} \operatorname{Gate}_z\left(\boldsymbol{x};\hat{\boldsymbol{\gamma}}_n\right) \operatorname{Expert}_z\left(y|\boldsymbol{x};\hat{\boldsymbol{\eta}}_{n,z}\right).\end{aligned} \tag{30}$$

Decision theoretic justification for the MAP and plugin-MAP rules for classification are detailed in McLachlan (1992).

## 4.2 Clustering

When conducting clustering, we assume that our data $\{\boldsymbol{D}_i\}_{i=1}^n$ arises from a DGP that can be best characterized by an MoE model, which is defined via the hierarchical construction that is characterized by Equations (1)–(3). That is, we assume that each $\boldsymbol{D}_i$ has a latent label $Z_i \in [g]$ that determines which of the $g$ experts of form (2) that it was generated from.

Let $\boldsymbol{d}$ be an arbitrary data point that is generated via a DGP that is best characterized by the $g_0$-component MoE (3) with some parameter vector $\boldsymbol{\theta}_0$. Suppose that we wish to estimate $Z$, the expert from which $\boldsymbol{d}$ was generated. As in Section 4.1, we can utilize a pair of MAP rule, depending on the nature of the clustering problem. The MAP rules for clustering are to estimate $Z$ by

$$\hat{z} = \arg\max_{z\in[g]} \frac{\operatorname{Gate}_z\left(\boldsymbol{x};\boldsymbol{\gamma}\right) \operatorname{Expert}_z\left(\boldsymbol{y}|\boldsymbol{x};\boldsymbol{\eta}_z\right)}{\operatorname{MoE}\left(\boldsymbol{y}|\boldsymbol{x};\boldsymbol{\theta}\right)}$$

which is equivalent to estimating $Z$ by

$$\hat{z} = \arg\max_{z\in[g]} \mathbb{P}\left(Z = z|\boldsymbol{X}_i = \boldsymbol{x}_i, \boldsymbol{Y}_i = \boldsymbol{y}_i\right),$$

or by

$$\hat{z} = \arg\max_{z \in [g]} \ \mathrm{Gate}_z\left(\boldsymbol{x}; \boldsymbol{\gamma}\right),$$

which is equivalent to

$$\hat{z} = \arg\max_{z \in [g]} \ \mathbb{P}\left(Z = z | \boldsymbol{X}_i = \boldsymbol{x}_i\right).$$

The estimation of the number of clusters experts, or clusters in such an application, can be conducted via the technique of Section 3.6. Furthermore, in general, we do not know the parameter vector $\boldsymbol{\theta}_0$ and thus we must estimate by the MQL estimator $\hat{\boldsymbol{\theta}}_n$ from a realization of $\{\boldsymbol{D}_i\}_{i=1}^{n}$. Using the estimated number of clusters (components) $\hat{g}_n$ and the MQL estimator for the MoE model with $\hat{g}_n$ clusters $\hat{\boldsymbol{\theta}}_n$, we obtain the plugin-MAP rules for clustering:

$$\hat{z} = \arg\max_{z \in [\hat{g}_n]} \ \frac{\mathrm{Gate}_z\left(\boldsymbol{x}; \hat{\boldsymbol{\gamma}}_n\right) \mathrm{Expert}_z\left(\boldsymbol{y} | \boldsymbol{x}; \hat{\boldsymbol{\eta}}_{n,z}\right)}{\mathrm{MoE}\left(\boldsymbol{y} | \boldsymbol{x}; \hat{\boldsymbol{\theta}}_n\right)} \tag{31}$$

and

$$\hat{z} = \arg\max_{z \in [\hat{g}_n]} \ \mathrm{Gate}_z\left(\boldsymbol{x}; \hat{\boldsymbol{\gamma}}_n\right). \tag{32}$$

## 4.3 Regression

Suppose now that we do not care that the DGP can be characterized via the hierarchical construction of Equations (1)–(3), but only that the conditional relationship between the response $\boldsymbol{Y}$ given input $\boldsymbol{X} = \boldsymbol{x}$ has form (3), for some parameter vector $\boldsymbol{\theta}_0$. We are often interested in using form (3) in order to estimate some functionals of the DGP of $\boldsymbol{d}$ such as the mean $\mathbb{E}\left(\boldsymbol{Y} | \boldsymbol{X} = \boldsymbol{x}\right)$ or higher moments. For example, in the case where the gating function is of form (4) and the expert arises from a location-scale family of conditional density functions, we can write $\mathbb{E}\left(\boldsymbol{Y}|\boldsymbol{X}=\boldsymbol{x}\right)$ in form (9).

Since the form of the MoE that best approximate the DGP of $\boldsymbol{d}$ is often unknown, we must estimate it via the realization of some random sample $\left\{\boldsymbol{D}_i\right\}_{i=1}^{n}$. Often both the number of components $g_0$ and the parameter vector $\boldsymbol{\theta}_0$ require estimation. As in Sections 4.1 and 4.2, $g_0$ can be estimated via $\hat{g}_n$, obtained via the technique from Section 3.6, and $\boldsymbol{\theta}_0$ can be estimated by the MQL estimator $\hat{\boldsymbol{\theta}}_n$.

Upon obtaining the estimators above, conditional moment functions can be easily computed. For example, if $\mathbb{E}\left[H\left(\boldsymbol{Y}\right)|\boldsymbol{X}=\boldsymbol{x}\right]$ exists, for some real-valued function $H\left(\boldsymbol{y}\right)$ of $\boldsymbol{y}\in\mathbb{Y}$, then

$$\mathbb{E}\left[H\left(\boldsymbol{Y}\right)|\boldsymbol{X}=\boldsymbol{x}\right]=\sum_{z=1}^{g_0}\text{Gate}_z\left(\boldsymbol{x};\boldsymbol{\gamma}_0\right)\int_{\mathbb{Y}}H\left(\boldsymbol{y}\right)\text{Expert}_z\left(\boldsymbol{y}|\boldsymbol{x};\boldsymbol{\eta}_{0,z}\right)\mathrm{d}\boldsymbol{y},\quad(33)$$

and can be estimated by

$$\hat{\mathbb{E}}\left[H\left(\boldsymbol{Y}\right)|\boldsymbol{X}=\boldsymbol{x}\right]=\sum_{z=1}^{\hat{g}_n}\text{Gate}_z\left(\boldsymbol{x};\hat{\boldsymbol{\gamma}}_n\right)\int_{\mathbb{Y}}H\left(\boldsymbol{y}\right)\text{Expert}_z\left(\boldsymbol{y}|\boldsymbol{x};\hat{\boldsymbol{\eta}}_{n,z}\right)\mathrm{d}\boldsymbol{y}.$$

We can obtain Equation (9) by making the substitution $H\left(Y\right)=Y$ into (33) for an MoE with soft-max gating functions and location-scale experts. If the conditional variance function of each expert is constant and equal to $\sigma_z^2$ for each $z\in[g]$, we can write the variance function of the response $Y$, given the input $\boldsymbol{X}=\boldsymbol{x}$, as

$$\text{var}\left(Y|\boldsymbol{X}=\boldsymbol{x}\right)=\sum_{z=1}^{g_0}\text{Gate}_z\left(\boldsymbol{x};\boldsymbol{\gamma}_0\right)\left[\left(\beta_{z,0}+\boldsymbol{\beta}_z^{\top}\boldsymbol{x}\right)^2+\sigma_z^2\right]-\left[\mathbb{E}\left(Y|\boldsymbol{X}=\boldsymbol{x}\right)\right]^2.$$

Furthermore, we can estimate $\text{var}\left(Y|\boldsymbol{X}=\boldsymbol{x}\right)$ by

$$\widehat{\text{var}}\left(Y|\boldsymbol{X}=\boldsymbol{x}\right)=\sum_{z=1}^{\hat{g}_n}\text{Gate}_z\left(\boldsymbol{x};\hat{\boldsymbol{\gamma}}_n\right)\left[\left(\hat{\beta}_{z,0}+\hat{\boldsymbol{\beta}}_z^{\top}\boldsymbol{x}\right)^2+\hat{\sigma}_z^2\right]-\left[\hat{\mathbb{E}}\left(Y|\boldsymbol{X}=\boldsymbol{x}\right)\right]^2.$$

Confidence intervals can also be constructed about any regression function. For example, see Nguyen & McLachlan (2014) regarding the construction of asymptotic confidence intervals around the mean function of an MoE model using the asymptotic normality conclusion of Theorem 6.

# 5 Example Applications

To demonstrate the applications of MoE models that are described in Section 4, we present the two following examples. We note that all computation for both examples are performed within the **R** programming environment (R Core Team, 2016) via the **flexmix** package of Grun & Leisch (2008), which allows for estimation of generic MoE models with soft-max gating functions. The optimization procedures utilized in **flexmix** are hybrid MM algorithms for MQL estimation and are described in Grun & Leisch (2008).

## 5.1 Three-Class Problem

We generate data $\{\boldsymbol{d}_i\}_{i=1}^n$ from the three-class problem of Chen et al. (1999) and Ng & McLachlan (2004). In the three-class problem, each data point $\boldsymbol{d}_i^\top = (\boldsymbol{x}_i^\top, y_i)$ consists of the input $\boldsymbol{x}_i = (x_{i1}, x_{i2})$, which is a realization of a random variable $\boldsymbol{X}_i^\top = (X_{i1}, X_{i2})$, where $X_{ij}$ is uniformly distributed over the interval $[-5, 5]$, for each $i \in [n]$ and $j \in \{1, 2\}$. Depending on $\boldsymbol{X}_i = \boldsymbol{x}_i$, the value of the response is a categorical variable $Y_i \in [3]$, such that $Y_i = 1$ by default, unless $\boldsymbol{x}_i$ is within a ball of radius two around the origin, in which case $Y_i = 2$, or if $\boldsymbol{x}_i$ is within the square with corners $(-4, 4)$ and $(-2, 2)$ or the square with corners $(2, 2)$ and $(4, 4)$, in which case $Y_i = 3$. A visualization of a realization of an $n = 1000$ observations sample from the three-class problem is provided in Figure 2. From the sample, we obtain 759, 156, and 85 observations, with category labels $y_i = 1, 2, 3$, respectively.

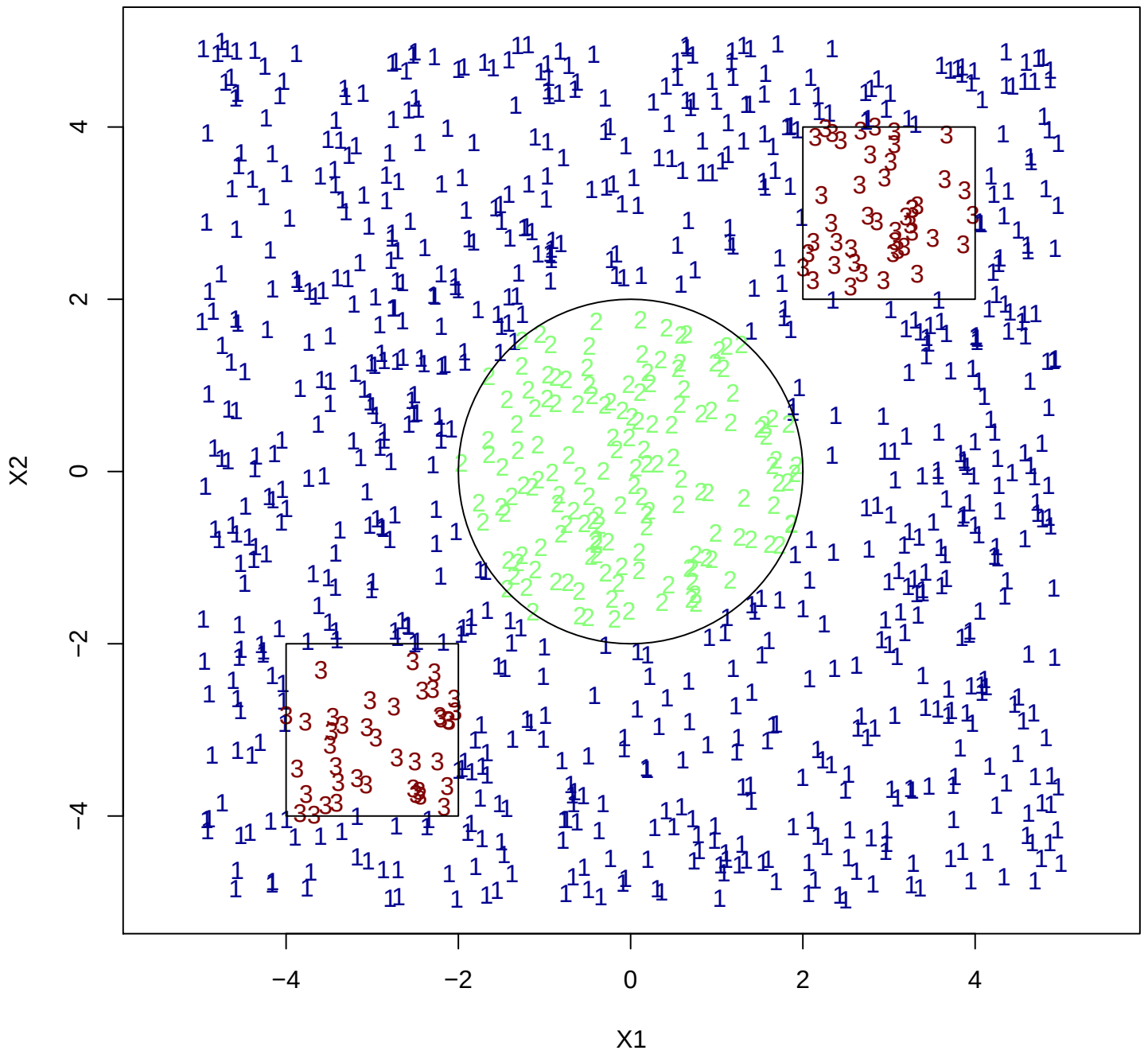

Figure 2: A realization of an $n = 1000$ observations sample from the three-class problem. The plot symbols indicate the class label.

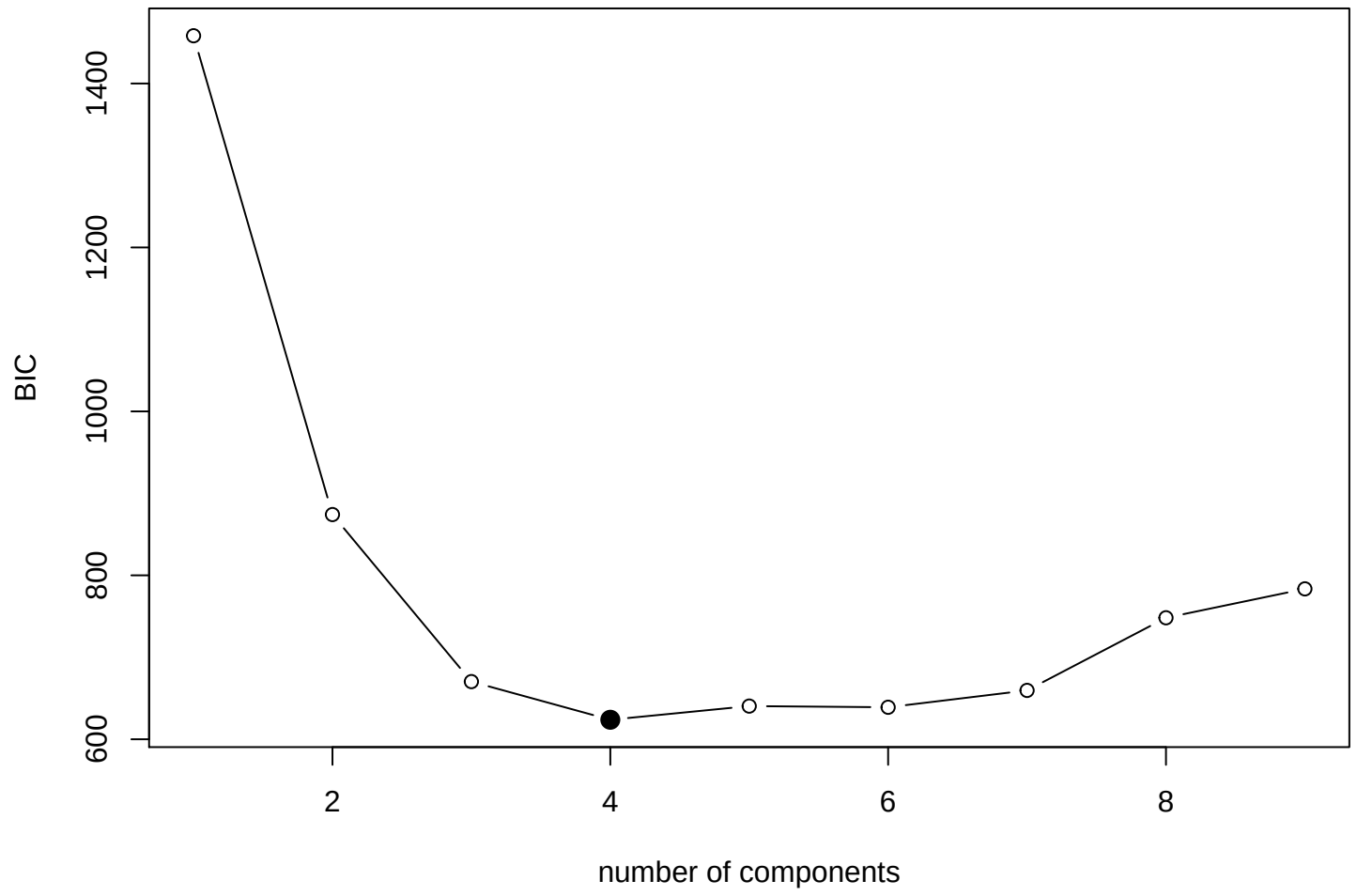

Figure 3: BIC values for $g \in [9]$ obtained from the MQL estimators on the $n = 1000$ observations sample from Figure 2. The filled marker indicates the best model obtained via rule (26).

In order to construct a classifier, based on the sample of $n = 1000$ observations from Figure 2, we estimate soft-max gated MoE models with multinomial logistic experts for $K = 3$ classes, as per Section 4.1, with varying numbers of components $g \in [9]$. The BIC values (i.e. twice the negative of the log-quasi-likelihood subtract the penalty of form (27)) for each $g$, obtained via MQL estimation, are plotted in Figure 3. From the figure, we observe that the best MoE model is that with $\hat{g}_n = 4$ components. Applying rule (30), we obtain a classification accuracy of 91.4% using the fitted MoE model classifier.

In order to better assess the performance of the MoE model classifier, defined by rule 30, we generate a new sample of $n = 2500$ observations from the sample process as described above. Using the fitted classifier, we obtain a test set accuracy rate (on the new sample) of 90.1%. A plot of the new sample along with the classifications via the MoE model classifier is displayed in Figure 4.

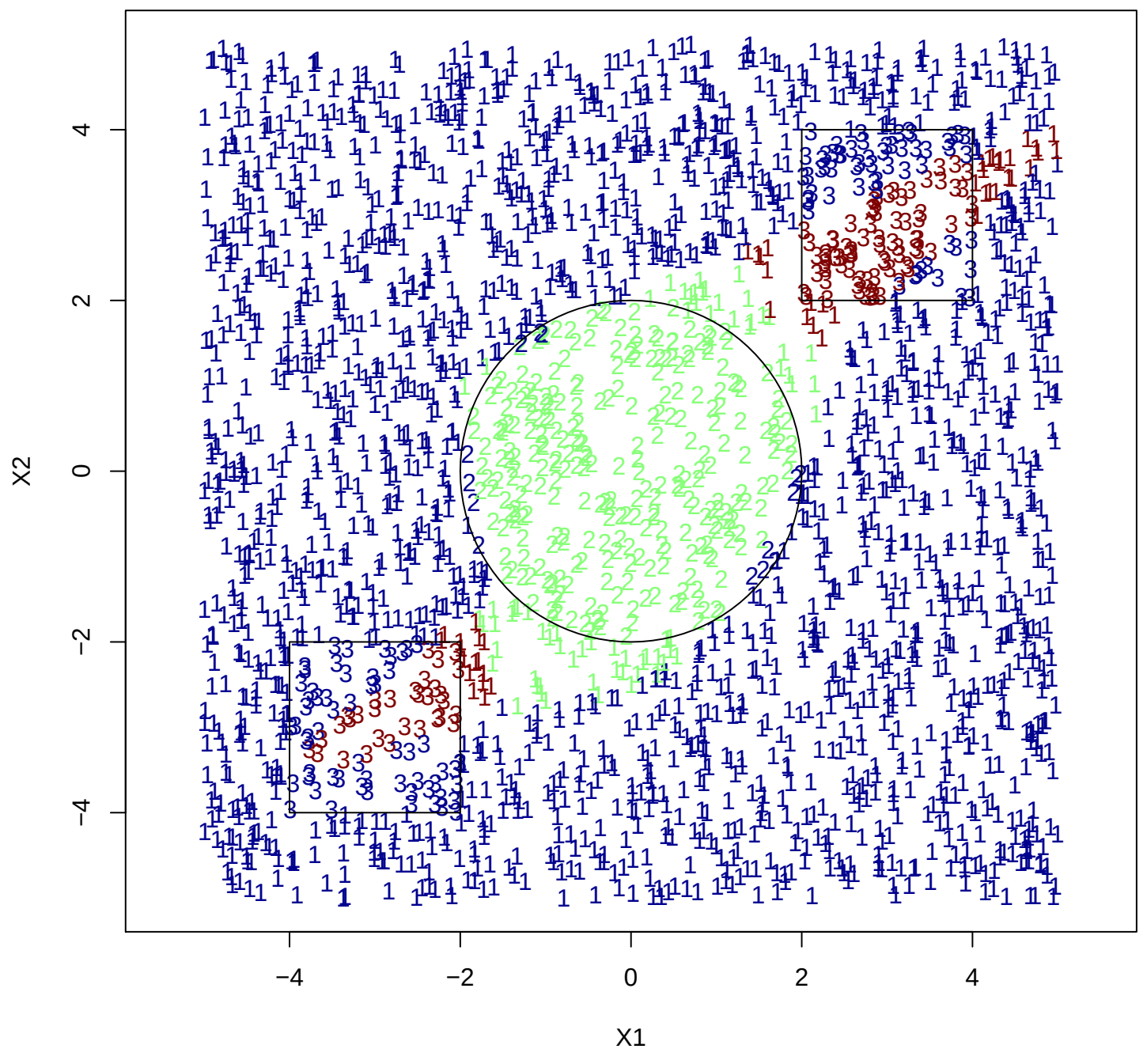

Figure 4: Plot of the additional $n = 2500$ observations sample from the three-class problem. The plot symbols indicate the true class labels $y_i$, $i \in [n]$. The color indicates the classification via the fitted MoE classifier, $\hat{y}_i$. Here, blue, green, and red correspond to $\hat{y}_i = 1, 2, 3$, respectively.

From Figure 4, we can obtain a visualization of the decision boundaries upon which the classification rule (30) assigns new point. We can see that the decision boundaries are not perfect fits to the rigid shapes of the true label boundaries of the DGP. However, upon inspection of Figure 2, we see that the classifier models the dense regions of each class very well. This is especially apparent when inspecting the dense regions of observations with $y_i = 3$ from the original sample. Also, where the classifier decision boundary exceeds the circle that determines when $y_i = 2$, we note that there are fewer observations and thus the lack of fit in that region of the domain is explainable. Considering that the percentage of observations in the training are 79.4%, 11.7%, and 8.9%, for $y_i = \{1, 2, 3\}$, respectively, the classification rate of 90.1% is a good result.

## 5.2 Switch Operation Power Signals

In this example, we analyze a time series data set arising from electrical signals at a switching point on the French railway, under a switching operation. The data were originally studied in Chamroukhi et al. (2009), Chamroukhi et al. (2010), and Same et al. (2011). An instance of such a signal, over a period of approximately 6 seconds, is presented in Figure 5. The signals are measured at $n = 550$ equally-spaced time points $x_i$ ($i \in [n]$) that are normalized to be in the unit interval. We let $y_i$ be the measurement of the power at each time point $x_i$, in Watts. Together $\{\boldsymbol{d}_i\}_{i=1}^n$ forms our sample of interest, where $\boldsymbol{d}_i^\top = (x_i, y_i)$ . We wish to model the power signals as a function of time.

Upon observation, it is clear that the time series in Figure 5 is highly nonlinear. Following the analysis by Chamroukhi et al. (2009), we model the DGP for the data via an MoE model with soft-max gating functions and quadratic-mean Gaussian regression experts of form

$$\text{Expert}_z\left(y|x;\boldsymbol{\eta}_z\right) = \phi\left(y;\beta_{0z} + \beta_{1z}x + \beta_{2z}x^2, \sigma_z^2\right),$$

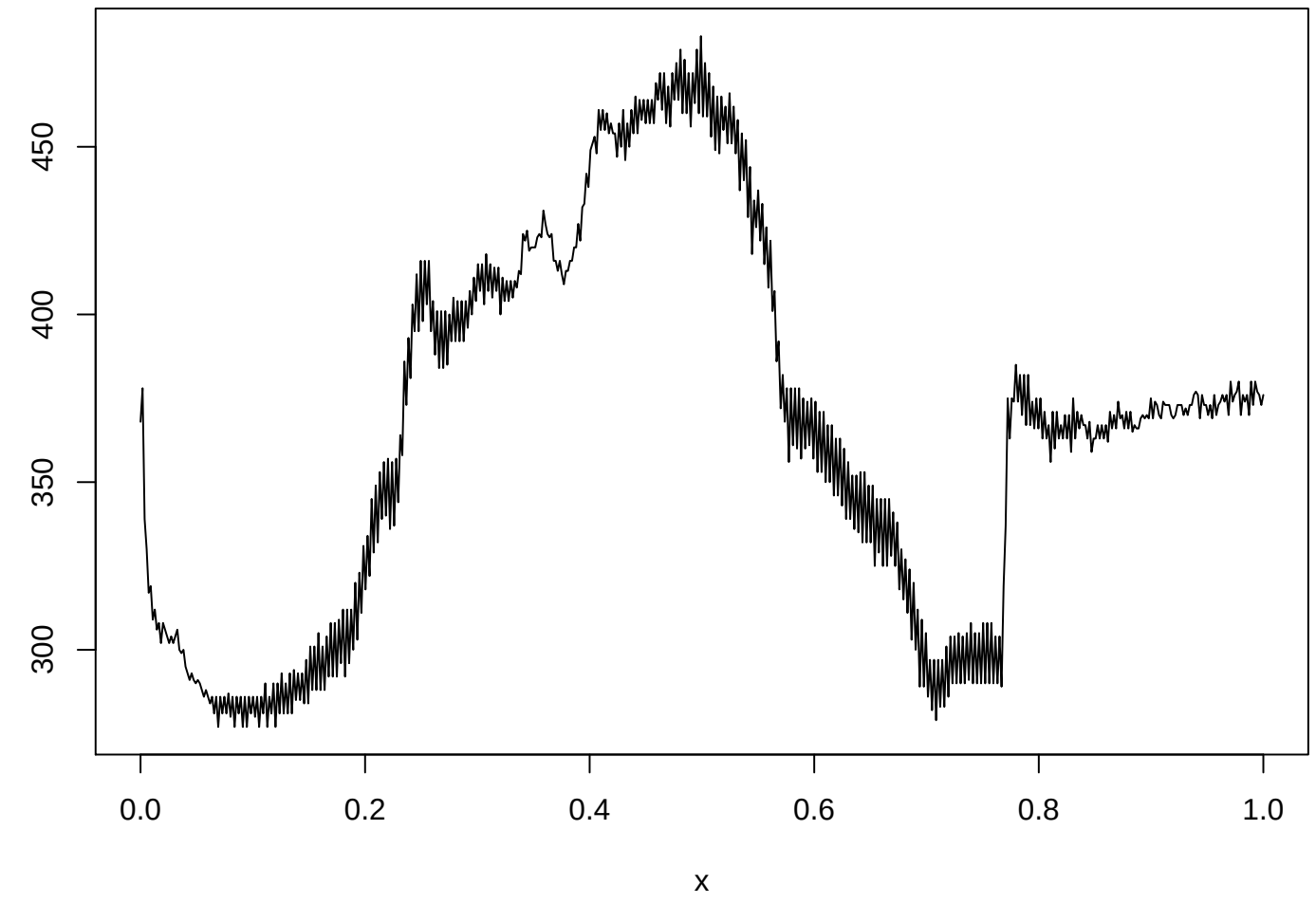

Figure 5: Instance of an electrical signal at a switching point, undergoing a switching operation. The abscissa displays the time at which the signal is measured (normalized to the unit interval) and the ordinate displays the value of the signal, in Watts.

where $\boldsymbol{\eta}_z^\top = \left(\beta_{0z}, \beta_{1z}, \beta_{2z}, \sigma_z^2\right)$, for $z \in [g]$. Figure 6 displays the BIC for each $g \in [10]$, obtained via MQL estimation. Using the optimal model with $\hat{g}_n = 8$, we can obtain the MoE model that best approximates the DGP for the data, and the expectation curve for the MoE model, of form (9), is plotted in Figure (7). We observe that that the curve is a good fit for the data and models its primary features, without being so specific as to model its idiosyncrasies.

As the data arises from an electrical control of a Railway switching point, it undergoes multiple stages of control. Each of the $\hat{g}_n = 8$ components from the obtained MoE model can be seen as one of these stages or sub-stages. We can utilize the clustering rule (32) in order to assign each time point to one of these stages. Figure 8 displays the segments of the time series that are assigned to each of the $\hat{g}_n = 8$ components, along with the mean curve corresponding to the respective component. We utilize clustering rule (32) instead of rule (31) as the different stages of control under a signal switching are entirely time-based. Here, the modeling of the power curves is incidental in identifying the different stages of control, in time. Rule (31) is more appropriate when clustering data that arise from multiple functions, with respect to time, that are pooled together.

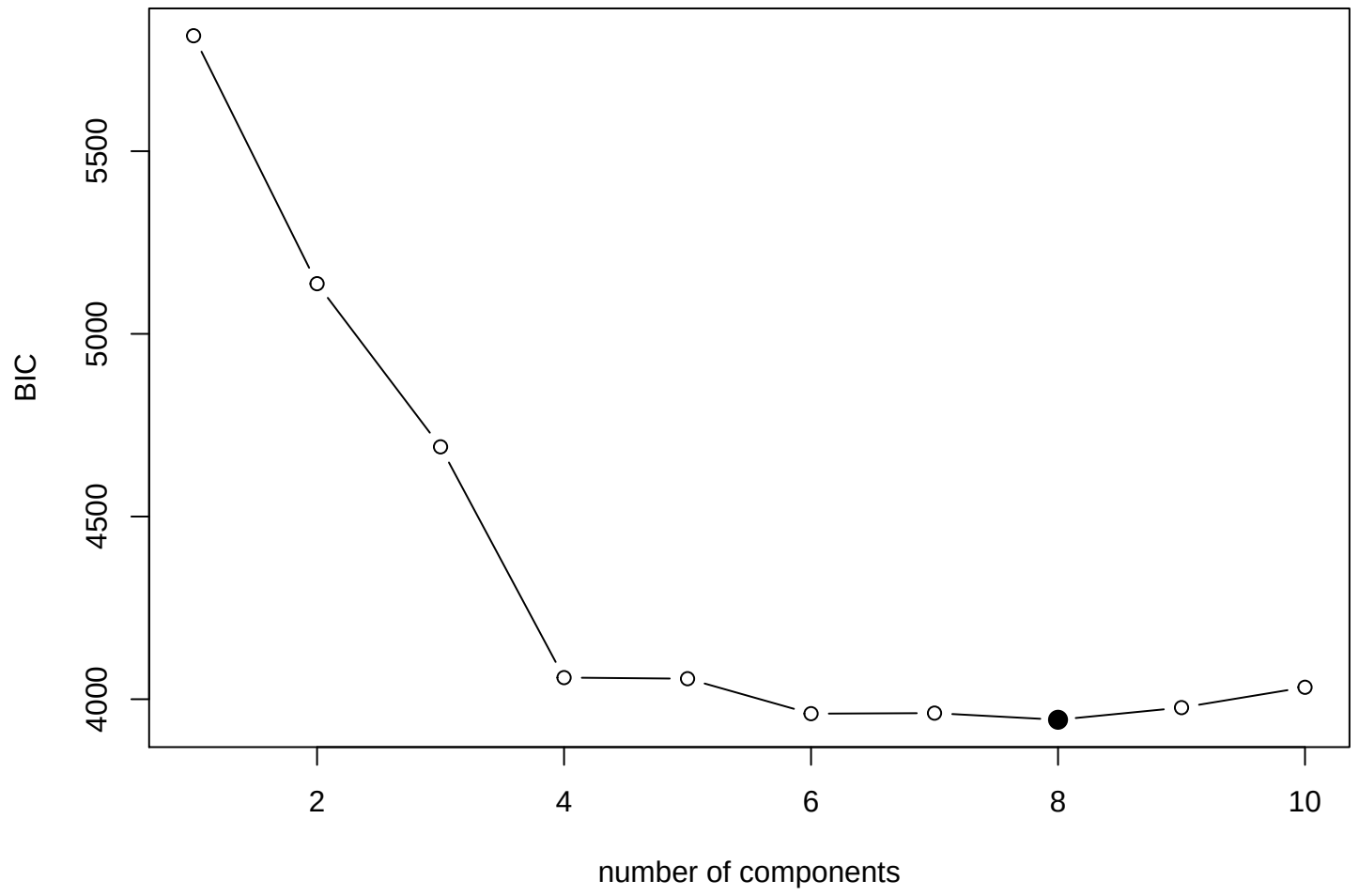

Figure 6: BIC values for $g \in [9]$ obtained from the MQL estimators on the $n = 550$ observations sample from the time series that is displayed in Figure 5. The filled marker indicates the best model obtained via rule (26).

# 6 Conclusions

MoE modeling is a powerful paradigm for approximating unknown DGPs, and for conducting classification, clustering, and regression. We have demonstrated how MoE models can be constructed for different data types, and we have provided theoretical results regarding the accuracy by which MoE models can approximate arbitrary DGPs and their mean functions.

When faced with data from an unknown DGP, MQL estimation can be used to estimate MoE models that are best fitted to the data in question. We have demonstrated that MQL estimation for MoE models can be conducted via blockwise-MM algorithms, and we have provided conditions under which the MQL estimators are consistent and asymptotically normal. A generic theorem is also provided for the construction of arbitrary information criteria for selecting the number of components of an MoE model.

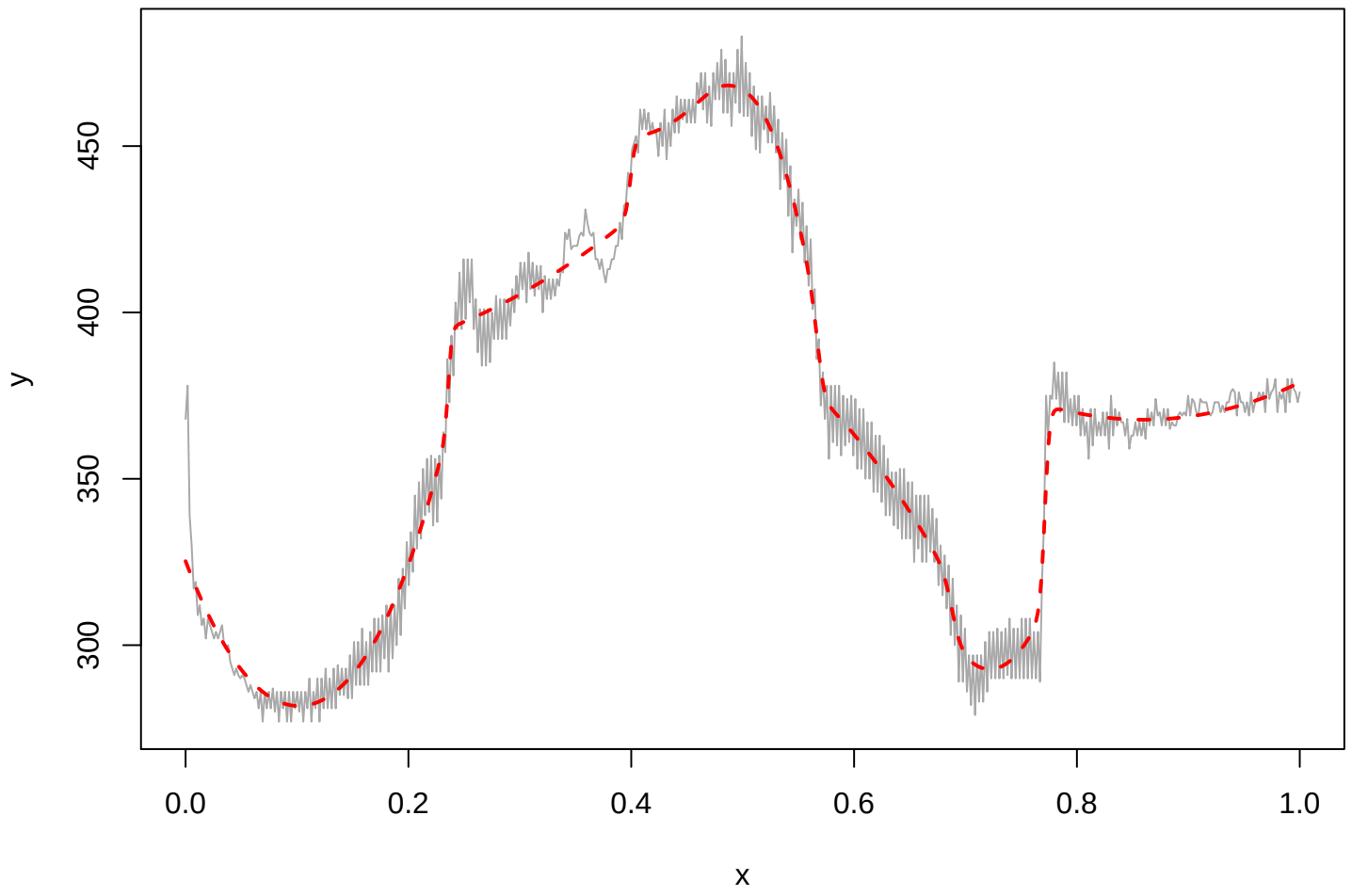

Figure 7: The original signal is plotted as a solid curve and the fitted mean function for the $\hat{g}_n = 8$ component MoE model, of form (9), is plotted as a dotted curve. The abscissa displays the time at which the signal is measured (normalized to the unit interval) and the ordinate displays the value of the signal, in Watts.

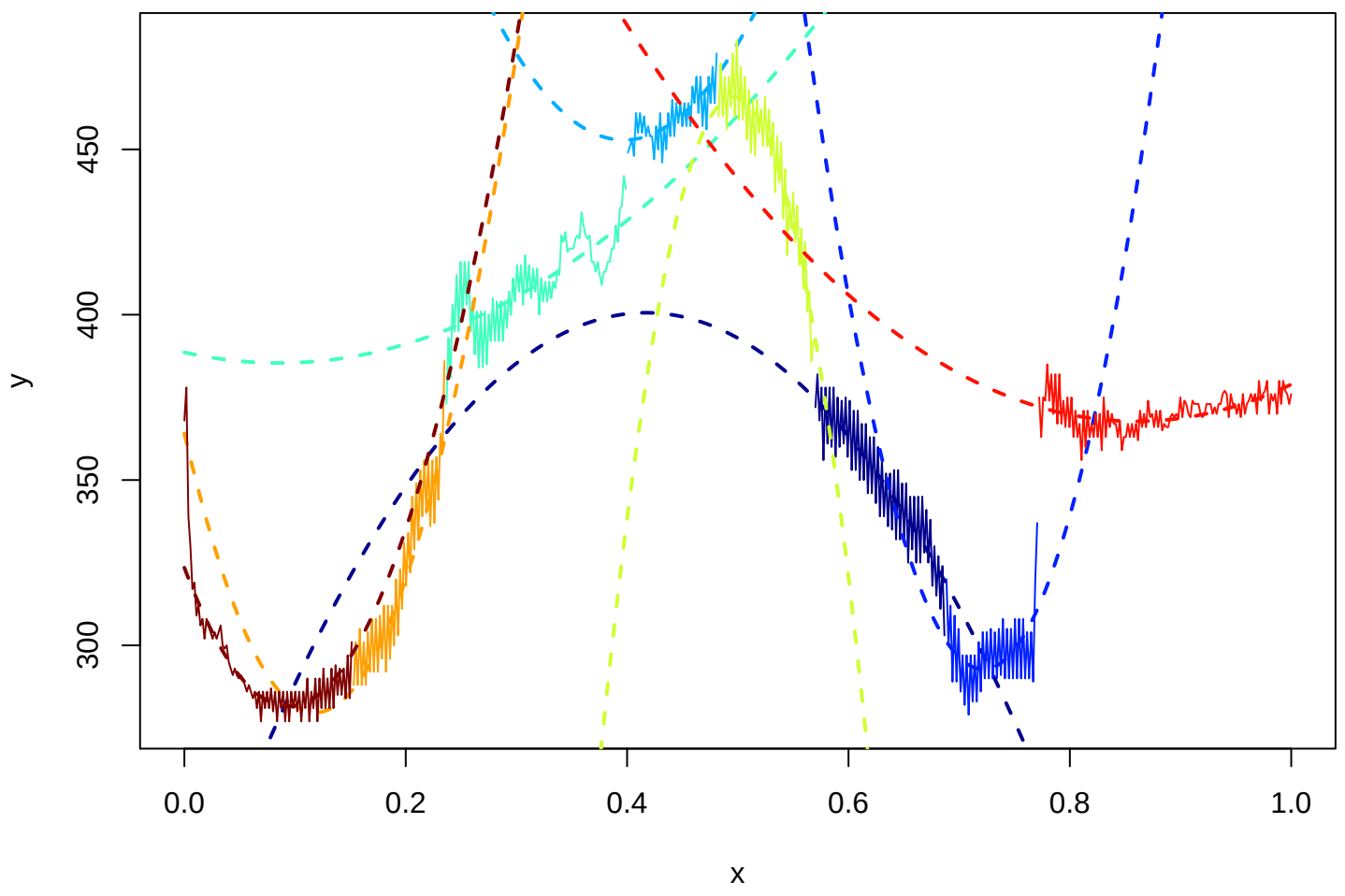

Figure 8: Each of the segments are colored in one of $\hat{g}_n = 8$ colors, and are plotted as solid curves. The dotted curves visualize the mean function of the corresponding MoE model component that the segment is clustered to. The abscissa displays the time at which the signal is measured (normalized to the unit interval) and the ordinate displays the value of the signal, in Watts.

To demonstrate their usefulness, we provide details regarding the conduct of classification, clustering, and regression via MoE models. A pair of examples has been provided to illustrate how these modes of application can be conducted in practice.

As with any review, summary, or tutorial article, we have omitted some details and topics for the sake of brevity and flow. A set of interest topics that we have omitted are works regarding the application of MoE models to the modeling of stationary time series processes, variable selection in MoE models via regularization, and finite sample model selection in MoE models. For the interested reader, we provide some details regarding these topics and the relevant literature, below.

The use of MoE for stationary time series modeling was first explored by Zeevi et al. (1999) who considered MoE models with soft-max gating functions and Gaussian autoregressive experts. The model of Zeevi et al. (1999) was further investigated in Carvalho & Tanner (2005a) alongside generic autoregressive expert functions. The family of MoE models with autoregressive GLM experts are explored in Carvalho & Tanner (2005b). A detailed investigation of the MoE model with autoregressive Poisson experts appears in Carvalho & Tanner (2007). A robust model using autoregressive Student-$t$ experts is considered in Carvalho & Skoulakis (2010). The use of hierarchical MoE models for modeling of univariate and multivariate time series process are investigated in Huerta et al. (2003) and Prado et al. (2006), respectively. Recent applications of Gaussian-gated MoE models for univariate and multivariate time series process appear in Kalliovirta et al. (2015) and Kalliovirta et al. (2016), respectively.

In our exposition, we have left out details regarding variable selection in MoE models with regression experts due to the topic being overly specific and because we cannot do it justice within the confines of this article. There has been a lot of recent interest in the topic of sparse variable selection via model regularization, that extend upon the pioneering work of Tibshirani (1996) and Fan & Li (2001). In the context of mixture modeling, studies regarding the performance of such estimators under various assumptions on underling DGPs can be found in Khalili & Chen (2007), Stadler et al. (2010), and Khalili & Lin (2013). Extensions of these regularization results to MoE models remain a recent area of interest and appear in Khalili (2010), Peralta & Soto (2014), and Shohoudi et al. (2016).

Lastly, we note that the information criteria approach from Section (3.6) is not the only available paradigm for choosing the number of components in an MoE model. Recent works by Cohen & Le Pennec (2014) and Montuelle & Le Pennec (2014) have demonstrated that the finite-sample variable selection approach of Massart (2007) can be adapted for use in the MoE context. Unfortunately, both works are limited to model selection for MoE models with Gaussian regression experts, only. An interesting future direction is to extend these works to construct model selection rules for general MoE models.